\documentclass[runningheads]{llncs}
\usepackage[T1]{fontenc}
\usepackage{graphicx}
\usepackage{amsmath}
\usepackage{natbib}
\usepackage{amssymb}
\usepackage{multirow}
\DeclareMathOperator*{\argmin}{argmin}
\usepackage{etoolbox}
\patchcmd{\thebibliography}{\section*}{\clearpage\section*}{}{}
\begin{document}
\title{Mitigating Text Toxicity with Counterfactual Generation}
%
%

\author{Milan Bhan\inst{1,2} \and Jean-Noel Vittaut\inst{2} \and Nina Achache\inst{1}, Victor Legrand\inst{1}, Annabelle Blangero\inst{1,3} \and Nicolas Chesneau\inst{1} \and Juliette Murris\inst{4} \and Marie-Jeanne Lesot\inst{2}} 

\institute{Ekimetrics \and
LFI, LIP6, Sorbonne Université, Paris, France \and
Aix-Marseille Université, Aix-Marseille, France \and
Université Paris Cité, Paris, France
}

\authorrunning{Bhan et al.}
%
%
\maketitle              
\begin{abstract}
Toxicity mitigation consists in rephrasing text in order to remove offensive or harmful meaning. Neural natural language processing (NLP) models have been widely used to target and mitigate textual toxicity. However, existing methods fail to detoxify text while preserving the initial non-toxic meaning at the same time. In this work, we propose to apply eXplainable AI (XAI) methods to both target and mitigate textual toxicity. We propose \texttt{CF-Detox}$_{\text{tigtec}}$ to perform text detoxification by applying local feature importance, counterfactual example generation and counterfactual feature importance methods to a toxicity classifier distinguishing between toxic and non-toxic texts. We carry out text detoxification through counterfactual generation on three datasets and compare our approach to three competitors. Automatic and human evaluations show that recently developed NLP counterfactual generators lead to competitive results in toxicity mitigation. This work is the first to bridge the gap between counterfactual generation and text detoxification and paves the way towards more practical applications of XAI methods.

\end{abstract}

\section{Introduction}
\label{sec:intro}
Online textual toxicity can be considered as rude, aggressive and degrading attitudes exhibited on online platforms, ranging from harmful to hateful speech. Hateful speech is defined as aggressive or offensive language against a specific group of people who share common characteristics, such as religion, race, gender, sexual orientation, sex or political affiliation~\cite{castano2021internet}. Such toxic content has proliferated on the Internet in the recent years~\cite{hate_online}, raising concerns about its multi-faceted negative impact, such as the potential to threaten the psychological and physical well-being of victims~\cite{social_media_hate_impact} or to be used as a medium for criminal actions~\cite{call_to_genocide}.

Toxic text data can also have a negative impact when used to train large language models (LLMs). Recent advances in natural language processing (NLP) and the development of LLMs such as GPT-3~\cite{brown_language_2020} have been made possible by utilizing vast quantities of textual data available on the Internet. These models have demonstrated a high capacity to generate plausible text, while raising several risks about harmful content generation~\cite{bender_dangers_2021} and bias amplification~\cite{gallegos_bias_2023} coming from the training texts. Thus, by generating toxic content, LLMs may contribute to the rapid spread of online toxicity, as as text is increasingly generated synthetically by chatbots~\cite{gen_ai_internet}.

\begin{figure}[t]{\centering}
\begin{center}
\includegraphics[scale=0.40]{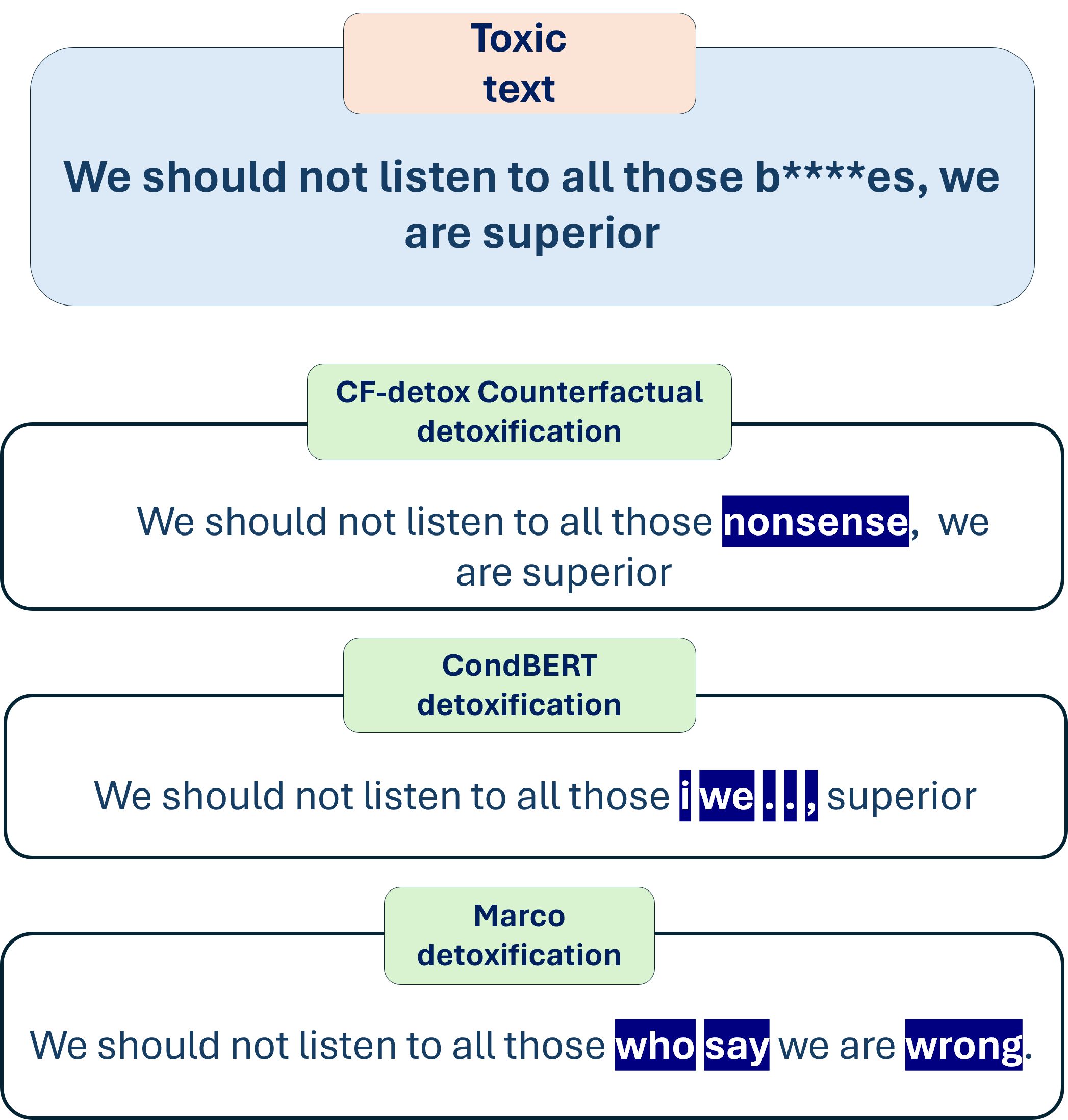}
\caption{Example of one text detoxification through counterfactual generation with our proposed \texttt{CF-Detox}$_{\text{tigtec}}$ compared to \texttt{MaRCo}~\cite{hallinan_detoxifying_2023} and \texttt{CondBERT}~\cite{dale_text_2021}. Text changes to mitigate toxicity are highlighted in blue. Explicitly toxic words have been censored with *.}
\label{fig:intro_fig}
\end{center}
\end{figure}

To cope with the rapid development of online toxic content and curb its societal impact, automatic toxicity processing methods have been developed to detect and process hateful content in online communities and digital media platforms~\cite{toxicity_survey_detection}. For example, such tools can be used to clean up language model training datasets, or to suggest interventions on toxic and hateful messages to the human-labor moderating online platforms. In particular, text detoxification (or toxicity mitigation) aims to rewrite toxic text in order to remove (or to mitigate) toxicity while preserving the initial non-toxic meaning and maintaining grammatical plausibility. Recent methods based on neural NLP models have been developed to perform text detoxification~\cite{hallinan_detoxifying_2023, dale_text_2021, laugier2021civil} by either generating text under constraint, or by detecting toxic content and modifying it. While these methods succeed in drastically lowering textual toxicity, they generally fail to preserve the initial non-toxic content. Besides, automatic online toxicity processing tools raise major ethical questions regarding the risks related to their robustness and their relationship with online human-labor.

In this paper we propose to address toxicity targeting and mitigation by applying eXplainable AI (XAI) methods~\cite{molnar_interpretable_2020}: more precisely Local Feature Importance (LFI)~\cite{barredo_arrieta_explainable_2020}, counterfactual (CF) generators~(see \cite{guidotti_counterfactual_2022} for a survey) and Counterfactual Feature Importance (CFI)~\cite{cfi}. The former aims at detecting important input features to explain a model prediction. Instead, CF example generation explains a model prediction by identifying the minimal changes that enable flipping the outcome of a classifier. CFI measures token modification importance to explain label flipping related to CF explanations. 

The main contributions of this work are as follows: \begin{enumerate}
    \item We show that LFI and CF generation can be applied to a toxicity classifier to target toxicity and to perform toxicity mitigation respectively,
    \item We propose \texttt{CF-Detox}$_{\text{tigtec}}$, a detoxification method based on a recently developed CF generator: \texttt{TIGTEC}~\cite{tigtec_2} and validated by both automatic and human experiments,
    \item We illustrate how to apply CFI methods to make detoxified texts even more content preserving,
    \item We discuss risks and opportunities related to automatic toxicity processing tools and define recommendations.
\end{enumerate}

For illustrative purposes, Figure~\ref{fig:intro_fig} shows an initial toxic text and its detoxified versions obtained from our proposed method, \texttt{MaRCo}~\cite{hallinan_detoxifying_2023} and \texttt{CondBERT}~\cite{dale_text_2021} respectively. Here, counterfactual detoxification leads to more plausible, sparse and context preserving text as compared to the other methods. 

The paper is organized as follows, in Section~\ref{sec:background_and_rw} we first recall basic principles about automatic toxicity processing and XAI. In Section~\ref{sec:detox_xai} we propose methods to use common XAI approaches to (1) target toxicity and (2) generate plausible and content-preserving detoxified texts. Experimental results discussed in Section~\ref{sec:xp} highlight that text detoxification through CF generation achieves competitive results in terms of toxicity mitigation, content preservation and plausibility, as compared to state-of-the-art competitors. This section also compares the ability of different LFI methods to target toxic content. Experimental evaluation is performed, both automatically and with a human-grounded protocol. After discussion and conclusion, we finally discuss in Section~\ref{sec:use_misuse} risks and opportunities around the use of toxicity mitigation methods. As a result, we discuss how to manage the diversity of toxicity perception and the risk of malicious use of detoxification tools and favor human-in-the-loop processes.

\section{Background and Related Work} \label{sec:background_and_rw}
This section recalls the task of automatic toxicity detection and mitigation. We presents existing methods aiming to detoxify text using neural NLP models. We then introduce the XAI principles used in the next section to perform toxicity detection and mitigation. In the following, we employ \textit{text detoxification} and \textit{toxicity mitigation} interchangeably.

\subsection{Automatic Toxicity Processing background}
\subsubsection{Definition and objective.}
Textual toxicity can be defined in multiple ways~\cite{toxicity_survey_detection} and can take various forms, such as rude, offensive or hateful speech, potentially causing online harm to isolated people, minority groups or different ethnic, religious or racial groups~\cite{hate_online}. Automatic toxicity processing can essentially take two forms: detection and mitigation. \textbf{Toxicity detection} can be either based on prior knowledge (vocabulary, regex) or obtained from a fine tuned toxicity classifier $f : \mathcal{X} \rightarrow \mathcal{Y}$ mapping an input text representation space $\mathcal{X}$ to an output space~$\mathcal{Y}$ to distinguish toxic and non-toxic texts~\cite{protecting_children}. Training language models to perform this classification is difficult, as it requires access to datasets labeled based on an idiosyncratic definition of toxicity derived from human annotators~\cite{jigsaw_18}. \textbf{Toxicity mitigation} consists in rewriting a toxic text while preserving the non-toxic meaning. This task is even more difficult because it requires disentangling toxic and non-toxic meanings to modify the former plausibly while preserving the latter.

\subsubsection{Evaluation criteria of text detoxification.}
Numerous desirable properties have been proposed to assess automatically text detoxification. We organize them into three categories. When ground truth detoxified texts are unavailable, we use the previously introduced $f$ classifier as an oracle to evaluate the toxicity level of the supposedly detoxified text. 
\emph{Accuracy} (\textbf{ACC}) measures the extent to which the generated texts are accurately detoxified with respect to~$f$. \emph{Accuracy} can be measured by computing either the rate of successful changes or the average toxicity score of the generated texts using~$f$. \emph{Proximity} or \emph{content preservation} (\textbf{CP}) evaluates how close two texts are. Textual similarity can be defined in two ways. The first one consists in evaluating textual proximity based on word sequence co-occurrences, with metrics such as self-BLEU~\cite{BLEU} or the Levenshtein distance. The second consists in measuring semantic similarity from word-level embeddings~\cite{glove} or sentence-level embeddings~\cite{zhang2019bertscore}. Finally, text \emph{plausibility} is automatically evaluated by the perplexity (\textbf{PPL}) score usually obtained from generative language models such as GPT-2~\cite{radford_language_nodate}.

\subsection{Toxicity Mitigation with Neural NLP Models}
This section describes two categories of methods leveraging neural NLP models to perform detoxification, namely Text Style Transfer (TST) and Masking and Reconstructing (M\&R).

\subsubsection{Text Style Transfer.} TST (see~\cite{tst_review} for a survey) aims to alter the stylistic attributes of an initial text while preserving its content that is unrelated to the target style. Text detoxification can be achieved through TST, where the initial style is characterized by the presence of toxicity, and the target style is defined by its absence. TST is usually performed by generating text with neural NLP decoders guided by an NLP toxicity classifier. In general, TST methods vary based on the language model used for autoregressive text generation and the heuristic of text generation steering. 

A first TST approach~\cite{detox_style_transfer} uses an encoder-decoder architecture based on recurrent neural networks to generate non-toxic text, using a toxicity convolutional neural network classifier to steer the style transfer. Another method~\cite{laugier2021civil} fine-tunes a text-to-text T5 model~\cite{rael_exploring_nodate} by using a denoising and cyclic auto-encoder loss. \texttt{ParaGeDi}~\cite{dale_text_2021} uses a pre-trained T5-based paraphraser model and a class-conditioned language model to steer the text generation. TST methods generally detoxify text accurately but struggle to preserve its non-toxic meaning~\cite{hallinan_detoxifying_2023}.

\subsubsection{Masking and Reconstructing.}
Masking and Reconstructing (M\&R) approaches perform toxicity mitigation by sequentially (1) targeting toxic content, (2) masking it, and (3) modifying it. Once the toxic content is targeted, mask infilling is usually performed with a neural NLP encoder. In general, M\&R methods differ in the way they target toxic content, and the neural NLP model used to perform mask infilling.

A first M\&R approach~\cite{detox_style_transfer_2} detoxifies text by retrieving potential harmful Part-Of-Speech (POS) based on a predefined toxicity vocabulary, generating non-offensive POS substitution candidates, and editing the initial text through mask infilling with a RoBERTa encoder for unacceptable candidates. \texttt{CondBERT}~\cite{dale_text_2021} identifies tokens to be masked using a logistic bag-of-words classifier and performs mask infilling using a BERT encoder. The most recent M\&R method called \texttt{MaRCo}~\cite{hallinan_detoxifying_2023} detects POS that could convey toxic meaning by comparing likelihoods from two BART encoder-decoders respectively fine-tuned on toxic and non-toxic content. The targeted potential toxic content is then replaced by non-toxic content by mixing token probabilities from these two encoder-decoders and a third neutral model. On average, M\&R yields to better results than TST in terms of content preservation, and performs equally regarding toxicity mitigation~\cite{hallinan_detoxifying_2023}.

\subsection{XAI for NLP}
In the following, we consider the neural NLP toxicity classifier $f : \mathcal{X} \rightarrow \mathcal{Y}$ introduced in the previous section, and a text $x = [t_{1},...,t_{d}] \in \mathcal{X}$ represented as a sequence of tokens of length $d$ with $f(x) = y$. $\mathcal{Y}$ can either be a binary space that distinguishes toxic and non-toxic texts or a multi-class space that categorizes several levels of toxicity.
\subsubsection{Local Feature Importance.}
A Local Feature Importance (LFI) function $g : \mathcal{X} \rightarrow \mathbb{R}^{d}$ explains a prediction by a vector $[z_{1},...,z_{d}]$ where $z_{i}$ is the contribution of the $i-$th token to the prediction. The higher the contribution, the more important the token is to explain the prediction of the classifier $f$. Three types of LFI methods can be distinguished: \textit{perturbation-based} such as KernelSHAP~\cite{lundberg_unified_2017}, \textit{gradient-based} such as Integrated gradients~\cite{sundararajan_axiomatic_2017} and \textit{attention-based} such as self-attention in case of a Transformer classifier~\cite{bhan_evaluating_2023}. 

\subsubsection{Counterfactual explanations.}
Counterfactual (CF) explanations emphasize what should be different in an input instance to change the associated outcome of a classifier~\cite{ali_explainable_2023}. CF examples provide contrastive explanations by simulating alternative instances to assess whether a specific event (in our case the predicted class) still occurs or not~\cite{miller_explanation_2019}. The CF example generation can be formalized as a constrained optimization problem. For a given classifier $f$ and an instance of interest $x$, a CF example $x_{\text{cf}}$ must be close to $x$ but predicted differently. It is defined as:
\begin{equation}
\label{eqn:argmin2}
   x_{\text{cf}} = \argmin_{z \in \mathcal{X}} c(x,z)  \: \text{ s.t. } \:  f(z) \neq f(x)  
\end{equation}
where $c : \mathcal{X} \times \mathcal{X} \rightarrow \mathbb{R}$ is a cost function that aggregates several expected CF characteristics, such as the textual distance. The CF explanation is then the difference between the generated CF example and the initial data point, $x_{\text{cf}} - x$. Many desirable characteristics for CF explanations have been proposed~\cite{guidotti_counterfactual_2022}, such as sparsity or plausibility to make sure that the CF example is not out-of-distribution~\cite{laugel_dangers_2019}. In Section~\ref{sec:detox_xai}, we present several methods to generate textual CFs by comparing and relating them with the toxicity mitigation task.

\subsubsection{Counterfactual Feature Importance.} A Counterfactual Feature Importance (CFI) operator $h : \mathcal{X} \times \mathcal{X} \rightarrow \mathbb{R}^{d}$ computes token substitution importance between an instance of interest and its related CF explanation~\cite{cfi}. In the same way as common LFI, $h_{i}$ is the contribution of the $i$-th token modification. CFI highlights the most important token modifications, making non-sparse CF explanations easier to read and more intelligible. A simple but effective way to compute CFI is to apply Integrated gradients to  explain the difference between $x_{cf}$ and $x$, by setting the baseline to the initial instance of interest $x$.

\section{When XAI meets text detoxification}  \label{sec:detox_xai}


In this section we show how LFI methods can foster toxic content targeting and discuss how to apply CF generation methods for performing text detoxification. We also propose a new method based on CFI to make detoxified texts even more content preserving.

\subsection{Targeting Toxic POS with Local Feature Importance}

Toxic POS targeting consists in identifying the elements in a toxic text that induce its toxicity. While toxicity can be easily defined partially by a predefined lexical field, it can also take on more complex forms implying specific cultural references or sarcasm that are difficult to detect automatically. Let $f$ be a toxicity classifier and $x$ is a toxic text with $f(x)=\texttt{toxic}$. 

We propose to apply LFI to~$f$ to highlight important tokens that explain why $x$ has been classified as toxic by $f$, enabling toxic POS detection. This way, toxic POS detection with LFI does not require the definition of a predefined toxicity vocabulary and is only based on a model that is trained to discriminate between toxic and non-toxic texts. Then, toxicity is detected in a \textit{data-driven} manner based on a fine-tuned neural NLP model. Our proposition to apply LFI methods to $f$ makes it possible to detect complex forms of toxicity, as recent neural NLP models such as BERT can encode high level linguistic forms to make their predictions. Toxic POS detection through LFI depends on the ability of the $f$ classifier to accurately discriminate between toxic and non-toxic texts. Consequently, a toxicity classifier with a low accuracy might misclassify toxic and non-toxic texts, leading to unreliable explanations.

\begin{figure}[t]{\centering}
\begin{center}
\includegraphics[scale=0.35]{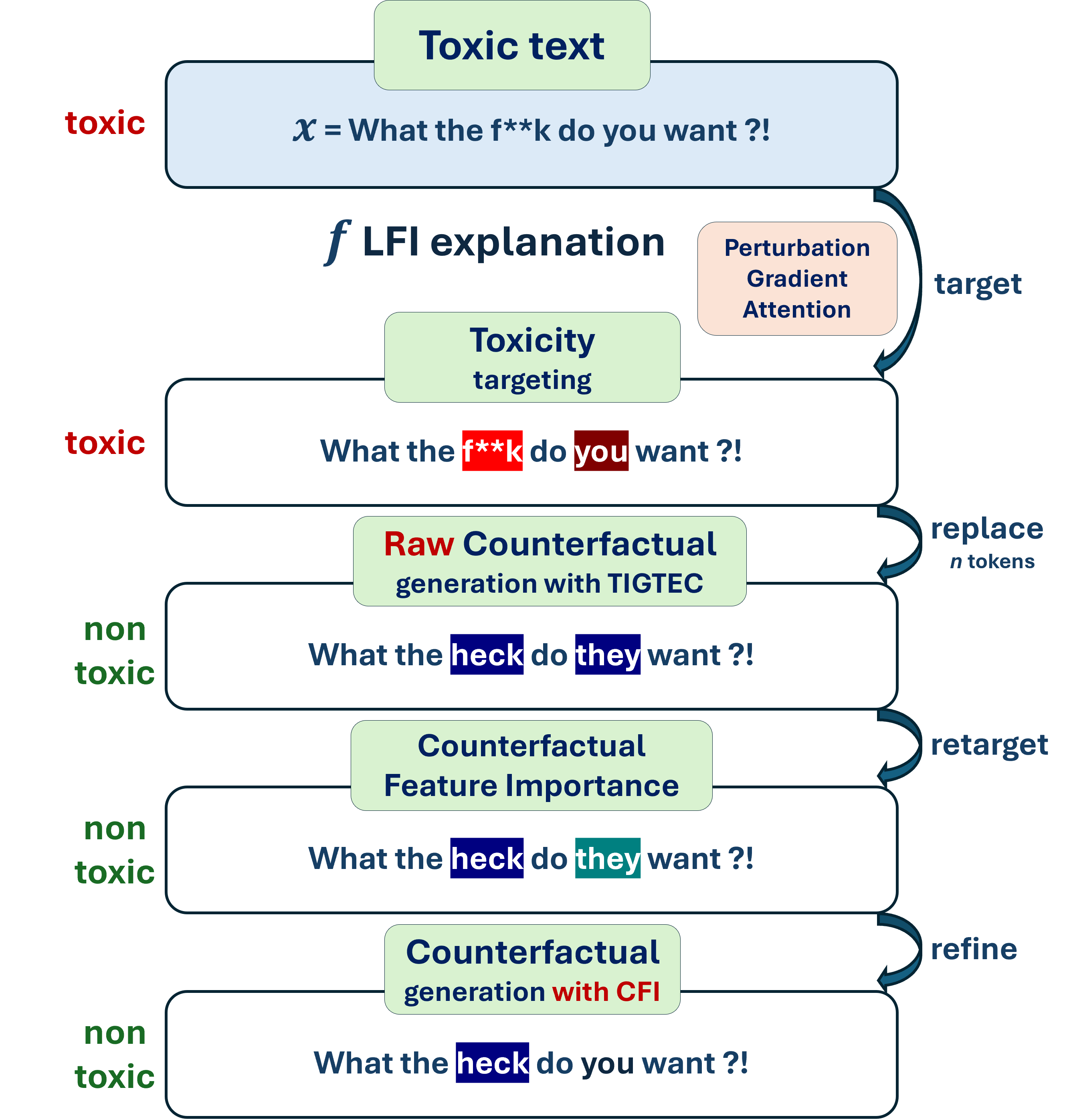}
\caption{Illustrative example of a toxic text and a CF detoxification process with our proposed \textit{target-then-replace-then-refine} approach. (1) Toxic content (in red) is targeted and (2) then modified (in blue). To be more content preserving, the counterfactual example is finally refined (3) by restarting the counterfactual generation process, guided by counterfactual feature importance. Explicitly toxic words have been censored with *. The darker the shade of color (red for LFI, blue for CFI), the higher the importance.}
\label{fig:cf_detox}
\end{center}
\end{figure}

The top part of Figure~\ref{fig:cf_detox} (toxicity targeting) shows an example of toxic POS targeting with LFI applied to a toxicity classifier: the token "\textbf{f**k}" is assessed as important to predict toxicity. In Section~\ref{sec:xp}, we experimentally study the relevance of this approach by comparing \emph{perturbation-based}, \emph{attention-based} and \emph{gradient-based} LFI methods.

\begin{figure*}[h]
    \centering
    \includegraphics[width=1\linewidth]{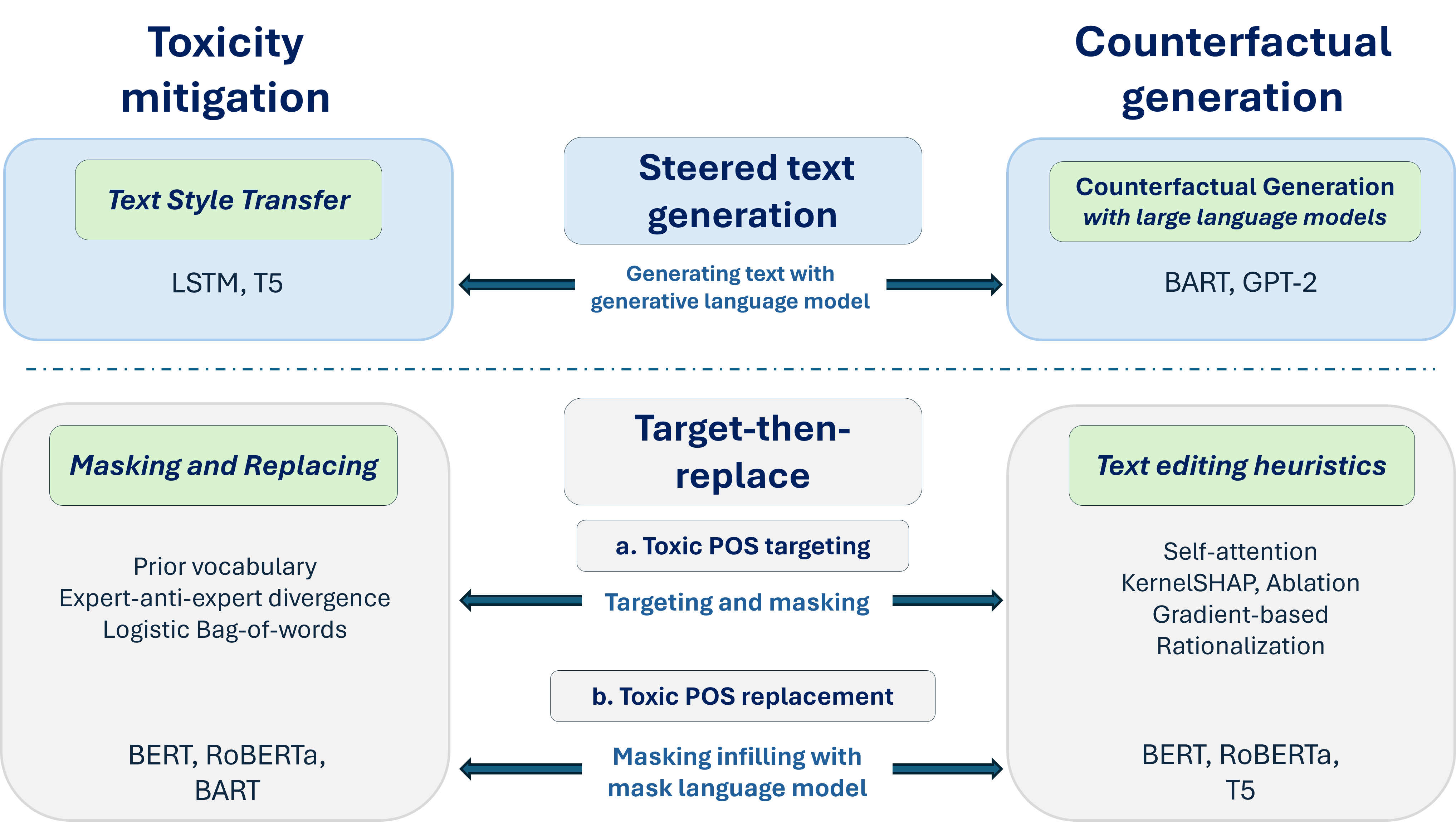}
    \caption{Toxicity mitigation and counterfactual generation comparison by method category. Toxicity mitigation methods and counterfactual generators can be categorized as \textit{steered text generation} and \textit{target-then-replace} approaches. Neural NLP models used to generate text or replace tokens are similar or of the same nature.}
    \label{fig:enter-label}
\end{figure*}

\subsection{Epistemic Similarities Between Text Detoxification and Counterfactual Generation}
In the following, we disclose that toxicity mitigation and CF generation share common features in terms of objective, evaluation criteria and categories of approaches.

\subsubsection{Similarity of objectives and evaluation criteria.}
Firstly, the aim of both toxicity mitigation and CF generation is to generate text close to an initial instance, while reaching a target state. For detoxification purpose, the target state is defined as non-toxicity, whereas the target state is defined by the $f$ classifier label for CF generation. Secondly, as discussed in Section~\ref{sec:background_and_rw}, toxicity mitigation methods share with CF generators several evaluation criteria, such as \emph{accuracy}, \emph{proximity} and \emph{plausibility}. These evaluation criteria are measured with the same metrics, such as sparsity, embedding-based semantic similarity, BLEU score or perplexity measures.     

\subsubsection{Similarity of constitutive categories of methods.}
Thirdly, methods detoxifying text and generating CFs can be grouped into two similar categories. As mentioned in~\cite{tigtec_2}, textual CF generators can be of two types: \emph{Text editing heuristics} and \emph{Counterfactual generation with large language models}.

\emph{Text editing heuristics} is a family of CF generation methods that builds CF examples by slightly modifying the input text whose prediction is to be explained. Important tokens are targeted and modified with mask language models to switch the outcome of the classifier, making this approach very similar to the M\&R way of detoxifying text. Text editing heuristics methods differ in the way they target important tokens and the language model used to modify the initial text. Regarding the former methods, they mostly target tokens to be modified by applying LFI methods to the classifier that has to be explained. For example, \texttt{CLOSS}~\cite{fern_text_2021} applies Ablation, \texttt{MiCE}~\cite{ross_explaining_2021} \textit{gradient-based} approaches, \texttt{CREST}~\cite{treviso_crest_2023} leverages rationalization methods, and \texttt{TIGTEC}~\cite{tigtec_2} employs KernelSHAP or Self-attention. Next, mask language models used to perform mask infilling are essentially T5, RoBERTa or BERT. This way, M\&R methods and Text editing heuristics differ only in the way they target toxic POS. As mentioned in the previous section, the former detect tokens to change either by the use of prior knowledge (vocabulary), intrinsically interpretable model (logistic model) or expert-anti-expert disagreement, whereas Text editing heuristics apply LFI methods to a toxicity classifier. Finally, text is modified using the same kind of NLP encoder. We propose to group these two kinds of approaches together under the name of "\textbf{target-then-replace}" methods. Figure~\ref{fig:cf_detox} shows the raw \textit{target-then-replace} CF detoxification process, where the detoxification consists in performing the two following token changes: \textbf{\fbox{f**k $\rightarrow$ heck}} and \textbf{\fbox{you $\rightarrow$ they}}, leading to a text certainly detoxified, but not perfectly content preserving.

\emph{Counterfactual generation with large language models} (CF-LLM) approaches build CF examples by leveraging pre-trained generative language models in the same way as TST text detoxifiers.
These methods differ in the language model applied to generate text and the heuristic used to steer the model towards a specific objective. For example, \texttt{CASPer}~\cite{madaan_plug_2022} learns perturbations to steer text generation with BART~\cite{lewis_bart_2020} and \texttt{Polyjuice} fine-tunes GPT-2 to generate CF examples. Thus, TST and CF-LLM methods differ in the way generative language models are steered towards a specific \textit{style} or \textit{label}. We group these two families of approaches together under the name of "\textbf{steered text generation}" methods. Figure~\ref{fig:enter-label} summarizes the connection between TST text detoxification and 
CF-LLM generation on the one hand, and M\&R methods and Text editing heuristics on the other.

All in all, CF generation and text detoxification can be (1) defined in the same way with the objective to find a small change to reach a target state, (2) evaluated with common metrics and (3) categorized in two similar families of methods, namely \textit{steered text generation} and \textit{target-then-replace}.

\subsection{Counterfactual Toxicity Mitigation Methodology}
\label{sec:methodology}
We propose to perform toxicity mitigation with a \textit{target-then-replace} CF generator with respect to the $f$ toxicity classifier. We postulate that, following the notations of Equation~\ref{eqn:argmin2}, toxicity mitigation can be performed by setting $x$ as a toxic instance of interest with $f(x) = \texttt{toxic}$. Therefore, the objective is to find $x_{\text{cf}}$ that minimizes the cost function $c$ such that $f(x_{\text{cf}}) = \texttt{non-toxic}$. This way, by denoting $\mathcal{M}$ as a \textit{target-then-replace} CF generation method and $h$ the LFI method used to target important tokens to be modified, $x_{cf}$ is defined such as $x_{cf} = \mathcal{M}(x, h(x))$ . The lack of ground truth detoxified texts is then overcome through the use of $f$ as an oracle to guide text detoxification while keeping the non-toxic content. This way, text detoxification through CF generation consists in detecting texts classified as toxic by $f$ and generating their related detoxified CF examples.

We also propose to use CFI to refine CF explanations (detoxified texts) and make them sparser, i.e. more content preserving. Since CF generation methods can generate explanations involving unnecessary token modifications (see token “\textbf{you}” in Figure~\ref{fig:cf_detox}), we re-run the CF generator, guided by CFI to optimally target the tokens that have to be modified to perform detoxification. Formally, refining a CF example with a CFI operator consists in generating $\tilde{x}_{cf}$ such as $\tilde{x}_{cf} = \mathcal{M}(x, h({x}_{cf}, x))$ with $h$ a CFI operator. We categorize this methodology as a \textit{target-then-replace-then-refine} approach. The entire Figure~\ref{fig:cf_detox} traces the whole \textit{target-then-replace-then-refine} process of CF detoxification. 

\section{Experimental Settings} 
\label{sec:xp}

This section presents the experimental studies conducted across three datasets. We compare our approach performing CF toxicity mitigation named \texttt{CF-Detox}$_{\text{tigtec}}$ to three competitors: \texttt{MaRCo}, \texttt{CondBERT} and \texttt{ParaGeDi}.

\subsection{Experimental Protocol}
\subsubsection{Datasets.}
We perform toxicity mitigation on three toxicity datasets from \cite{hallinan_detoxifying_2023}. \textbf{Microagression.com} (MAgr) is a public blog containing socially-biased interactions with offending quotes.
\textbf{Social Bias Frames} (SBF) is
a corpus of offensive content from various online sources. We use a subset of SBF where the texts have been labeled as hateful by annotators. \textbf{DynaHate} is a dataset of hate comments that are difficult to detect for a hate-speech classifier. Toxicity mitigation is run on texts initially classified as toxic in all three cases. MAgr contains 425 toxic texts, SBF 557 and DynaHate 662. The three datasets are available in the Github project in the initial paper\footnote{\url{https://github.com/shallinan1/MarcoDetoxification/tree/main/datasets}}.

\subsubsection{Counterfactual generator and competitors.}
We instantiate our method proposed in the previous section by setting as CF generation method \texttt{TIGTEC}~\cite{tigtec_2}. The rationale behind this choice is that \texttt{TIGTEC} achieves the best compromise in terms of accuracy, content preservation and plausibility when compared with other major competitors~\cite{tigtec_2}. \texttt{TIGTEC} is a \emph{target-then-replace} textual CF generator that implements several LFI methods to target important tokens to be changed. It iteratively masks and replaces tokens with a BERT mask language model following a tree search policy based on beam search to minimize a cost function. We use the following settings to run \texttt{TIGTEC}, we first train a neural classifier (BERT or DistilBERT) on a toxic task dataset from Kaggle\footnote{\url{https://www.kaggle.com/datasets/rounak02/imported-data}} to learn to distinguish toxic texts from non-toxic ones. The classifier performance after training is respectively 94.1\% and 90.2\% for BERT and DistilBERT. This way, CF text detoxification is performed with classifiers that have been trained on a different dataset from the ones used for the evaluation. 

Toxicity mitigation is then performed by generating CF examples starting from toxic texts in order to reach a non-toxic state. In the following, we call \textbf{\texttt{CF-Detox}$_{\text{tigtec}}$} our new toxicity mitigation method based on CF generation. The cost function driving CF search is defined as in the TIGTEC paper as $c(x_{\text{cf}},x) = -\left(p(y_{\text{non-toxic}}|x_{\text{cf}}) - \alpha d_{s}(x_{\text{cf}}, x)\right)$ with $d_{s}(x,x') = \frac{1}{2}(1 - s(x,x'))$ where $p(y_{\text{non-toxic}})$ is the probability score of being non-toxic with respect to the trained classifier, $s$ is a cosine similarity measure between text embeddings obtained from Sentence Transformers and $\alpha$ is a hyperparameter set to 0.3. This way, minimizing the cost function implies reaching a counterfactual state while staying semanticly close to the initial instance. The TIGTEC counterfactual search is performed in a beam search fashion, with $\texttt{beam\_width}=4$. For each neural model, \textbf{\texttt{CF-Detox}$_{\text{tigtec}}$} is run in three different versions, targeting toxic POS with either KernelSHAP (kshap), Self-attention (attention) or Integrated gradients (IG). Self-attention is aggregated as in the original \texttt{TIGTEC} paper by averaging all the attention coefficients related to the CLS token over the attention heads in the last layer of the $f$ classifier. 

We compute CFI to refine the raw CF examples generated by \textbf{\texttt{CF-Detox}$_{\text{tigtec}}$} by explaining the differences between the initial instances and their raw detoxified versions via Integrated gradients. CFI is only applied to CF examples (raw detoxified texts) initially involving at least two token modifications. 

\textbf{\texttt{CF-Detox}$_{\text{tigtec}}$} is compared to the three identified most recent text toxicity mitigation methods : \textbf{\texttt{MaRCo}}~\cite{hallinan_detoxifying_2023}, \textbf{\texttt{CondBERT}} and \textbf{\texttt{ParaGeDi}}~\cite{dale_text_2021}. 

\subsubsection{Automatic evaluation.}
We use the 5 metrics previously introduced to assess toxicity mitigation. The toxicity metrics are based on a pre-trained toxicity classifier. The library used to import the pre-trained toxicity classifier is \verb|transformers| and the model backbone is \verb|toxic-bert|. This toxicity classifier is different from the one used to steer \texttt{CF-Detox}$_{\text{tigtec}}$ toxicity mitigation. The success rate is computed with the accuracy (\textbf{\%ACC}) from the binary classifier to assess if the evaluated text is toxic or non-toxic. \textbf{\%ACC} is defined as the number of non-toxic texts over the total number of evaluated texts, with respect to the pre-trained toxicity classifier. The average toxicity score (\textbf{SCORE}) is obtained from the last layer of the classifier. 

The sparsity (\textbf{\%S}) is computed with the normalized word-based Levenshtein distance. The content preservation (\textbf{\%CP}) is computed with the cosine similarity between Sentence Transformer~\cite{reimers_sentence-bert_2019} embeddings to evaluate the semantic proximity between the initial toxic text and its detoxified version. The library used to import the Sentence Transformer is \verb|sentence_transformers| and the model backbone is \verb|paraphrase-MiniLM-L6-v2|. 

Text plausibility is measured with the perplexity score \cite{jelinek_perplexitymeasure_2005} and compared to the perplexity of the original text ($\Delta\textbf{PPL}$). This way, a $\Delta$PPL score lower than 1 indicates than the text plausibility increases whereas $\Delta$PPL greater than 1 means that the detoxified text is less plausible. This score is computed based on the exponential average cross-entropy loss of Gemma-2B~\cite{team2024gemma}, a recently developed small generative language model outperforming GPT-2. The library to import the pre-trained model is \verb|transformers| and the backbone is \verb|gemma-2b|. Due to the presence of outliers when calculating the entropy used to calculate perplexity, aggregation is performed with the median rather than the mean operator.

\subsubsection{Human-grounded evaluation.}

In addition to automatic evaluation, we conduct a human-grounded experiment to compare \texttt{CF-Detox}$_{\text{tigtec}}$, \texttt{MaRCo}, \texttt{CondBERT} and \texttt{ParaGeDi} in terms of toxicity mitigation performance. It consists in asking 5 annotators to rank detoxified texts by toxicity level obtained by applying \texttt{CF-Detox}$_{\text{tigtec}}$, \texttt{MaRCo}, \texttt{CondBERT} and \texttt{ParaGeDi} on 20 randomly selected texts from each dataset. The order of appearance of the toxicity mitigation methods and the dataset is randomized, so that there is no spatial bias in information processing. Annotators can rank texts at the same level if necessary.

Before running the experiment, annotators are given the same instructions. To make sure that they annotate based on the same common knowledge, we define the textual toxicity as "\textit{violent, aggressive or offensive language that may focus on a specific person or group of people sharing a common property. This common property can be gender, sexual orientation, ethnicity, age, religion or political affiliation.}". Annotators all have a MSc degree in data analytics or machine learning and have a good knowledge of English. None of the annotators are authors. 


\subsection{Results}
\label{sec:result}
\subsubsection{Global results.}

\begin{table*}[t]
\centering
\caption{Counterfactual toxicity mitigation comparison to competitors on three datasets. Counterfactual toxicity mitigation is either based on a BERT or a DistilBERT classifier and target toxicity based on KernelSHAP.}
\small
\begin{tabular}{ccccccc|}
\hline
\multicolumn{1}{|c|}{\textbf{Dataset}}          & \multicolumn{1}{c|}{\textbf{Metric}}          & \multicolumn{1}{c|}{\textbf{\texttt{CondBert}}} & \multicolumn{1}{c|}{\textbf{\texttt{MaRCo}}} & \multicolumn{1}{c|}{\textbf{\texttt{ParaGeDi}}} & \multicolumn{2}{c|}{\textbf{\texttt{CF-Detox}$_{\text{tigtec}}$}} \\
\multicolumn{1}{|c|}{}                          & \multicolumn{1}{c|}{}                         & \multicolumn{1}{c|}{}                           & \multicolumn{1}{c|}{}                        & \multicolumn{1}{c|}{}                           & \textbf{Bert-kshap}           & \textbf{Distilbert-kshap}           \\ \hline
\multicolumn{1}{|c|}{\multirow{5}{*}{Dynahate}} & \multicolumn{1}{c|}{$\Delta$PPL $\downarrow$} & \multicolumn{1}{c|}{1.37}                       & \multicolumn{1}{c|}{0.70}                    & \multicolumn{1}{c|}{\textbf{0.43}}              & 1.24                         & 1.25                               \\ \cline{2-7} 
\multicolumn{1}{|c|}{}                          & \multicolumn{1}{c|}{\%CP $\uparrow$}          & \multicolumn{1}{c|}{68.8}                       & \multicolumn{1}{c|}{68.4}                    & \multicolumn{1}{c|}{71.2}                       & 86.4                         & \textbf{86.6}                      \\ \cline{2-7} 
\multicolumn{1}{|c|}{}                          & \multicolumn{1}{c|}{\%S $\uparrow$}           & \multicolumn{1}{c|}{64.1}                       & \multicolumn{1}{c|}{67.8}                    & \multicolumn{1}{c|}{31.1}                       & 86.1                         & \textbf{86.3}                      \\ \cline{2-7} 
\multicolumn{1}{|c|}{}                          & \multicolumn{1}{c|}{\%ACC $\uparrow$}         & \multicolumn{1}{c|}{\textbf{92.3}}              & \multicolumn{1}{c|}{70.2}                    & \multicolumn{1}{c|}{89.9}                       & 82.9                         & 81.9                               \\ \cline{2-7} 
\multicolumn{1}{|c|}{}                          & \multicolumn{1}{c|}{SCORE $\downarrow$}       & \multicolumn{1}{c|}{\textbf{0.10}}              & \multicolumn{1}{c|}{0.30}                    & \multicolumn{1}{c|}{0.14}                       & 0.21                         & 0.23                               \\ \hline
                                                &                                               &                                                 &                                              &                                                 &                              &                                    \\ \hline
\multicolumn{1}{|c|}{\multirow{5}{*}{MAgr}}     & \multicolumn{1}{c|}{$\Delta$PPL $\downarrow$} & \multicolumn{1}{c|}{2.59}                       & \multicolumn{1}{c|}{0.72}                    & \multicolumn{1}{c|}{\textbf{0.69}}              & 2.04                         & 2.16                               \\ \cline{2-7} 
\multicolumn{1}{|c|}{}                          & \multicolumn{1}{c|}{\%CP $\uparrow$}          & \multicolumn{1}{c|}{66.2}                       & \multicolumn{1}{c|}{65.2}                    & \multicolumn{1}{c|}{78.4}                       & 84.6                         & \textbf{86.1}                      \\ \cline{2-7} 
\multicolumn{1}{|c|}{}                          & \multicolumn{1}{c|}{\%S $\uparrow$}           & \multicolumn{1}{c|}{47.3}                       & \multicolumn{1}{c|}{66.0}                    & \multicolumn{1}{c|}{33.0}                       & \textbf{70.1}                & \textbf{70.1}                      \\ \cline{2-7} 
\multicolumn{1}{|c|}{}                          & \multicolumn{1}{c|}{\%ACC $\uparrow$}         & \multicolumn{1}{c|}{\textbf{98.1}}              & \multicolumn{1}{c|}{82.1}                    & \multicolumn{1}{c|}{91.6}                       & 95.1                         & 91.8                               \\ \cline{2-7} 
\multicolumn{1}{|c|}{}                          & \multicolumn{1}{c|}{SCORE $\downarrow$}       & \multicolumn{1}{c|}{\textbf{0.05}}              & \multicolumn{1}{c|}{0.20}                    & \multicolumn{1}{c|}{0.13}                       & 0.10                         & 0.14                               \\ \hline
                                                &                                               &                                                 &                                              &                                                 &                              &                                    \\ \hline
\multicolumn{1}{|c|}{\multirow{5}{*}{SBF}}      & \multicolumn{1}{c|}{$\Delta$PPL $\downarrow$} & \multicolumn{1}{c|}{1.48}                       & \multicolumn{1}{c|}{0.82}                    & \multicolumn{1}{c|}{\textbf{0.59}}              & 1.27                         & 1.26                               \\ \cline{2-7} 
\multicolumn{1}{|c|}{}                          & \multicolumn{1}{c|}{\%CP $\uparrow$}          & \multicolumn{1}{c|}{75.6}                       & \multicolumn{1}{c|}{74.7}                    & \multicolumn{1}{c|}{75.6}                       & 90.2                         & \textbf{90.4}                      \\ \cline{2-7} 
\multicolumn{1}{|c|}{}                          & \multicolumn{1}{c|}{\%S $\uparrow$}           & \multicolumn{1}{c|}{68.1}                       & \multicolumn{1}{c|}{68.1}                    & \multicolumn{1}{c|}{37.3}                       & 87.6                         & \textbf{88.4}                      \\ \cline{2-7} 
\multicolumn{1}{|c|}{}                          & \multicolumn{1}{c|}{\%ACC $\uparrow$}         & \multicolumn{1}{c|}{\textbf{95.9}}              & \multicolumn{1}{c|}{70.2}                    & \multicolumn{1}{c|}{91.7}                       & 87.6                         & 84.2                               \\ \cline{2-7} 
\multicolumn{1}{|c|}{}                          & \multicolumn{1}{c|}{SCORE $\downarrow$}       & \multicolumn{1}{c|}{\textbf{0.07}}              & \multicolumn{1}{c|}{0.31}                    & \multicolumn{1}{c|}{0.12}                       & 0.16                         & 0.21                               \\ \hline
\end{tabular}

\label{tab:results_1}
\end{table*}

Table~\ref{tab:results_1} shows the experimental results obtained by running each method on the same datasets. In this table, \texttt{CF-Detox}$_{\text{tigtec}}$ has been run using KernelSHAP to target toxic POS. For each dataset, \texttt{CF-Detox}$_{\text{tigtec}}$ leads to the most content preserving texts, with the highest \%CP and \%S scores. On the other hand, \texttt{CF-Detox}$_{\text{tigtec}}$ performs in average worse than \texttt{CondBERT} in terms of detoxification accuracy and score across all datasets. Still, the toxicity of texts generated by \texttt{CF-Detox}$_{\text{tigtec}}$ is in average lower than that of \texttt{MarCo} and similar to \texttt{ParaGeDi} over all text corpora. If \texttt{CondBERT} mitigates the most toxicity, the resulting detoxified texts are significantly different from the initial ones in terms of sparsity and semantic proximity. \texttt{CondBERT} generates the less plausible text across all datasets and degrades text plausibility whereas \texttt{ParaGedi} and \texttt{MaRCo} improve it. In particular, \texttt{ParaGeDi} produces the most plausible detoxified text. This result stems from the autoregressive properties of the paraphrase language model used by \texttt{ParaGeDi} to generate text that is intended to be plausible. However, the text generated by \texttt{ParaGeDi} diverges 
the most from the initial text in terms of sparsity and proximity.

The high perplexity level of the text generated by \texttt{CF-Detox}$_{\text{tigtec}}$ can be partially attributed to the mask language model used for generating new text. Indeed, it is significantly smaller than the encoder-decoder models used by \texttt{ParaGeDi} and \texttt{MaRCo} for text generation. The model used by \texttt{CF-Detox}$_{\text{tigtec}}$ is a \textit{small} 66M parameters DistilBERT for masked language model, whereas each encoder-decoder model used by \texttt{MaRCo} and \texttt{ParaGeDi} to generate text are respectively a 139M parameters BART and a 220M parameters T5. Using a bigger mask language model such as BERT-base or BERT-large would improve the plausibility of the text generated by \texttt{CF-Detox}$_{\text{tigtec}}$. \texttt{CF-Detox}$_{\text{tigtec}}$ still generates significantly more plausible texts as compared to \texttt{CondBERT}.

\begin{figure}[t]{\centering}
\begin{center}
\includegraphics[scale=0.32]{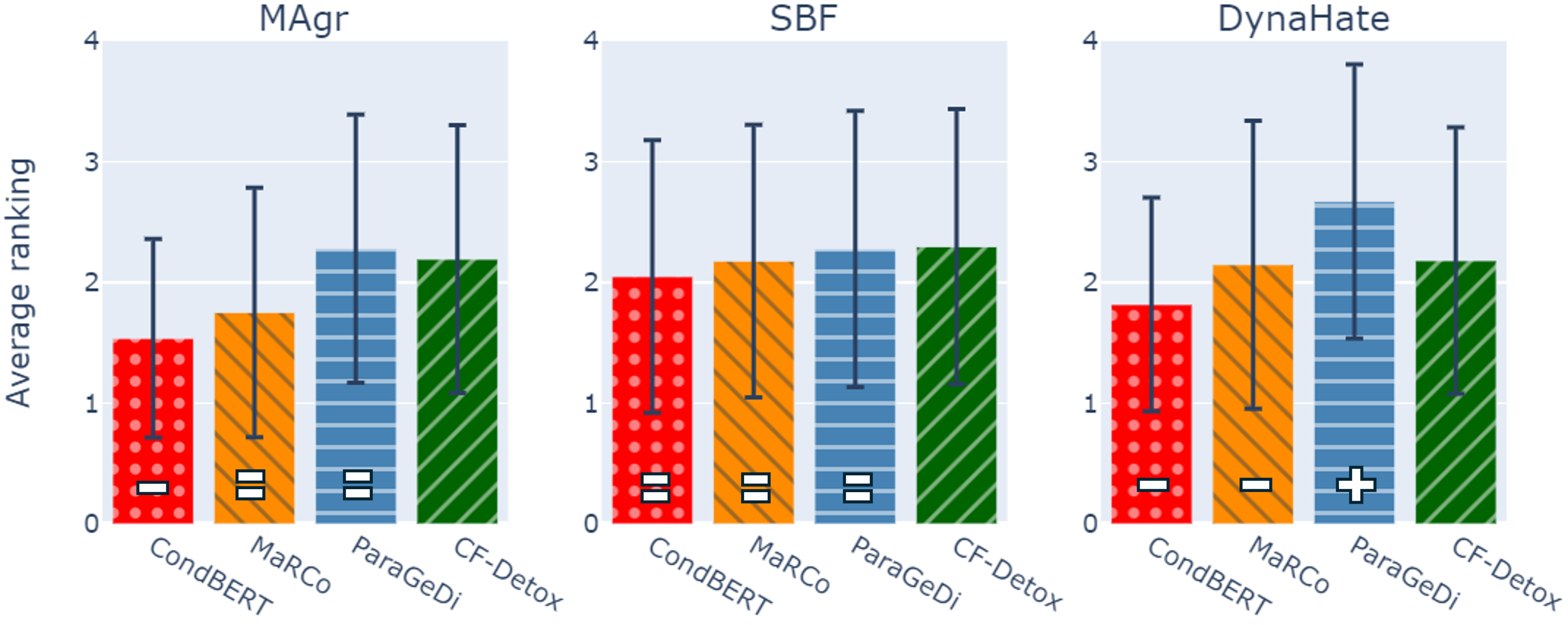}
\caption{Toxicity comparison on three test sets with a human-grounded experimental ranking evaluation. Competitor rank distributions are compared to \texttt{CF-Detox}$_{\text{tigtec}}$ using a one-tailed paired t-test with a 5\% threshold risk. The sign "-" indicates that the rank is lower in average as compared to \texttt{CF-Detox}$_{\text{tigtec}}$, whereas "=" and "+" respectively indicate that the ranking is in average similar and greater.}
\label{fig:xp_human}
\end{center}
\end{figure}

\subsubsection{Human evaluation.}

Figure~\ref{fig:xp_human} shows the results from the human-grounded experiment where human annotators rank methods' outputs by level of toxicity. \texttt{CondBERT} achieves the lowest level of toxicity on DynaHate and MAgr, which is consistent with the automatic analysis. \texttt{CF-Detox}$_{\text{tigtec}}$ and \texttt{MaRCo} produce less toxic texts as compared to \texttt{ParaGeDi} on the DynaHate dataset. Toxicity is overall at the same level across \texttt{CF-Detox}$_{\text{tigtec}}$, \texttt{MaRCo} and \texttt{ParaGeDi} on SBF.

This way, automatic and human evaluation underpin that \texttt{CF-Detox}$_{\text{tigtec}}$ offers another possible compromise between toxicity, meaning preservation and text plausibility as compared to other state-of-the-art existing methods.


\begin{table*}[t]
\centering
\caption{\texttt{CF-Detox}$_{\text{tigtec}}$ counterfactual toxicity mitigation by LFI POS toxicity targeting method.}
\small
\begin{tabular}{cccccccc}
\hline
\multicolumn{1}{|c|}{\textbf{Dataset}}          & \multicolumn{1}{c|}{\textbf{Metric}}          & \multicolumn{6}{c|}{\textbf{\texttt{CF-Detox}$_{\text{tigtec}}$}}                                                                       \\
\multicolumn{1}{|c|}{}                          & \multicolumn{1}{c|}{}                         & \multicolumn{3}{c|}{\textbf{Bert}}                                 & \multicolumn{3}{c|}{\textbf{Distilbert}}                           \\
\multicolumn{1}{|c|}{}                          & \multicolumn{1}{c|}{}                         & attention     & IG            & \multicolumn{1}{c|}{kshap}          & attention     & IG            & \multicolumn{1}{c|}{kshap}          \\ \hline
\multicolumn{1}{|c|}{\multirow{5}{*}{Dynahate}} & \multicolumn{1}{c|}{$\Delta$PPL $\downarrow$} & 1.34          & 1.26          & \multicolumn{1}{c|}{\textbf{1.25}} & 1.43          & \textbf{1.24} & \multicolumn{1}{c|}{1.25}          \\ \cline{2-8} 
\multicolumn{1}{|c|}{}                          & \multicolumn{1}{c|}{\%CP $\uparrow$}          & \textbf{87.4} & 86.3          & \multicolumn{1}{c|}{86.4}          & \textbf{89.2} & 87.0          & \multicolumn{1}{c|}{86.5}          \\ \cline{2-8} 
\multicolumn{1}{|c|}{}                          & \multicolumn{1}{c|}{\%S $\uparrow$}           & 86.8          & \textbf{87.4} & \multicolumn{1}{c|}{86.1}          & \textbf{87.4} & \textbf{87.4} & \multicolumn{1}{c|}{86.3}          \\ \cline{2-8} 
\multicolumn{1}{|c|}{}                          & \multicolumn{1}{c|}{\%ACC $\uparrow$}         & 81.6          & \textbf{84.0} & \multicolumn{1}{c|}{82.9}          & 79.0          & 81.4          & \multicolumn{1}{c|}{\textbf{81.9}} \\ \cline{2-8} 
\multicolumn{1}{|c|}{}                          & \multicolumn{1}{c|}{SCORE $\downarrow$}       & 0.23          & \textbf{0.20} & \multicolumn{1}{c|}{0.21}          & 0.27          & 0.23          & \multicolumn{1}{c|}{\textbf{0.23}} \\ \hline
                                                &                                               &               &               &                                    &               &               &                                    \\ \hline
\multicolumn{1}{|c|}{\multirow{5}{*}{MAgr}}     & \multicolumn{1}{c|}{$\Delta$PPL $\downarrow$} & 2.38          & 2.14          & \multicolumn{1}{c|}{\textbf{2.04}} & 2.76          & 2.26          & \multicolumn{1}{c|}{\textbf{2.16}} \\ \cline{2-8} 
\multicolumn{1}{|c|}{}                          & \multicolumn{1}{c|}{\%CP $\uparrow$}          & \textbf{88.0} & 85.8          & \multicolumn{1}{c|}{84.6}          & \textbf{90.5} & 87.1          & \multicolumn{1}{c|}{86.1}          \\ \cline{2-8} 
\multicolumn{1}{|c|}{}                          & \multicolumn{1}{c|}{\%S $\uparrow$}           & \textbf{70.7} & \textbf{70.7} & \multicolumn{1}{c|}{70.1}          & \textbf{71.4} & 70.8          & \multicolumn{1}{c|}{70.1}          \\ \cline{2-8} 
\multicolumn{1}{|c|}{}                          & \multicolumn{1}{c|}{\%ACC $\uparrow$}         & 94.4          & \textbf{95.5} & \multicolumn{1}{c|}{95.1}          & 89.0          & \textbf{92.0} & \multicolumn{1}{c|}{91.8}          \\ \cline{2-8} 
\multicolumn{1}{|c|}{}                          & \multicolumn{1}{c|}{SCORE $\downarrow$}       & \textbf{0.13} & 0.10          & \multicolumn{1}{c|}{0.10}          & \textbf{0.19} & 0.14          & \multicolumn{1}{c|}{0.14}          \\ \hline
                                                &                                               &               &               &                                    &               &               &                                    \\ \hline
\multicolumn{1}{|c|}{\multirow{5}{*}{SBF}}      & \multicolumn{1}{c|}{$\Delta$PPL $\downarrow$} & 1.35          & \textbf{1.26} & \multicolumn{1}{c|}{1.27}          & 1.32          & 1.27          & \multicolumn{1}{c|}{\textbf{1.26}} \\ \cline{2-8} 
\multicolumn{1}{|c|}{}                          & \multicolumn{1}{c|}{\%CP $\uparrow$}          & \textbf{91.5} & 90.4          & \multicolumn{1}{c|}{90.2}          & \textbf{92.3} & 90.9          & \multicolumn{1}{c|}{90.4}          \\ \cline{2-8} 
\multicolumn{1}{|c|}{}                          & \multicolumn{1}{c|}{\%S $\uparrow$}           & 88.7          & \textbf{89.2} & \multicolumn{1}{c|}{87.6}          & 89.0          & \textbf{89.3} & \multicolumn{1}{c|}{88.4}          \\ \cline{2-8} 
\multicolumn{1}{|c|}{}                          & \multicolumn{1}{c|}{\%ACC $\uparrow$}         & 86.4          & 86.5          & \multicolumn{1}{c|}{\textbf{87.6}} & 82.4          & 84.0          & \multicolumn{1}{c|}{\textbf{84.2}} \\ \cline{2-8} 
\multicolumn{1}{|c|}{}                          & \multicolumn{1}{c|}{SCORE $\downarrow$}       & 0.19          & 0.17          & \multicolumn{1}{c|}{\textbf{0.16}} & 0.24          & \textbf{0.21} & \multicolumn{1}{c|}{\textbf{0.21}} \\ \hline
\end{tabular}
\label{tab:results_2}
\end{table*}

\subsubsection{Ablation study.}

Table~\ref{tab:results_2} shows the experimental results obtained by running three different versions of \texttt{CF-Detox}$_{\text{tigtec}}$ on the three datasets of interest. Each \texttt{CF-Detox}$_{\text{tigtec}}$ instance is defined by the LFI method used to target toxic POS. Table~\ref{tab:results_2} shows that Self-attention, Integrated gradients and KernelSHAP lead to similar results in terms of toxicity mitigation, content preservation and text plausibility. These results highlight that LFI methods of a different nature (perturbation, attention or gradient-based) can all yield good results, which underpins \texttt{CF-Detox}$_{\text{tigtec}}$ robustness.

Toxicity mitigation through CF generation methods like \texttt{TIGTEC} has to be performed by choosing the appropriate LFI method to target toxicity based on the available model information. For example, KernelSHAP is appropriate if no information (internal parameters, gradients) is available about the classifier $f$ used to counterfactually mitigate toxicity, due to its \textit{model-agnostic} nature. On the contrary, if $f$ gradients are accessible, the use of Integrated gradients is recommended since it is less computationally costly than KernelSHAP. Finally, if all $f$ parameters are accessible, using Self-attention is appropriate because it is available at no cost.

Since \texttt{TIGTEC} gradually masks and replaces tokens in the original toxic text based on LFI, we postulate that the sparser detoxified texts a LFI method induces, the better its performance, as it targets the most discriminating tokens of the initial text with respect to $f$. This way, integrated gradients give the most \emph{faithful} explanations (i.e. target the most accurately toxicity).

\subsubsection{CFI enhancement.}

\begin{table*}[t]
\centering
\caption{CFI-induced sparsity and similarity gains.}
\begin{tabular}{|c|c|c|c|}
\hline
\textbf{Model}              & \textbf{Dataset} & \textbf{Similarity} & \textbf{Sparsity} \\ \hline
\multirow{3}{*}{BERT}       & MAgr             & +0.5\%               & +1.2\%             \\ \cline{2-4} 
                            & SBF              & +0.4\%               & +2.2\%             \\ \cline{2-4} 
                            & DynaHate         & \textbf{+1.5\%}      & \textbf{+4.7\%}    \\ \hline
\multirow{3}{*}{DistilBERT} & MAgr             & +0.3\%               & +0.9\%             \\ \cline{2-4} 
                            & SBF              & +0.5\%               & +2.2\%             \\ \cline{2-4} 
                            & DynaHate         & \textbf{+2.0\%}      & \textbf{+4.8\%}    \\ \hline
\end{tabular}
\label{tab:cfi_impact}
\end{table*}

\begin{figure*}[t]
    \centering
    \includegraphics[width=0.55\linewidth]{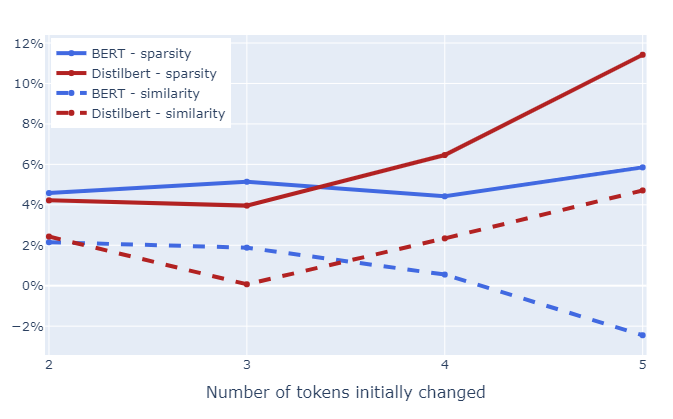}
    \caption{Evolution of the gain in sparsity and similarity induced by CFI on Dynahate, according to the number of tokens initially modified, per model.}
    \label{fig:evol_cfi_impact}
\end{figure*}

We refine detoxified texts using CFI following the \textit{target-then-replace-then-refine} methodology previously introduced in Section~\ref{sec:methodology}. Table~\ref{tab:cfi_impact} shows similarity and sparsity growth rate obtained from CFs obtained by running \texttt{CF-Detox}$_{\text{tigtec}}$ with KernelSHAP to target toxic POS.  On average, CFI increases both similarity and sparsity across all datasets, when \texttt{CF-Detox}$_{\text{tigtec}}$ is either based on BERT or DistilBERT. However, the gains are higher with the DynaHate dataset (respectively +4.7\% and +4.8\% uplift in sparsity for BERT and DistilBERT) containing the most toxic texts as compared to MAgr and SBF.

Figure~\ref{fig:evol_cfi_impact} displays the relation between the gain induced by CFI and the number of tokens initially modified in the first version of the CF by focusing on the DynaHate dataset. The number of considered token modifications stops at 5, representing 95\% of the CFs initially generated. It turns out that for DistilBERT, the higher the number of initially modified tokens, the higher the sparsity and proximity gain (up to +11\% for sparsity). Results are overall more stable for BERT across the number of tokens initially changed. 
\section{Discussion}

In this work we showed that XAI methods can be applied to a toxicity classifier to target toxic POS with LFI and mitigate toxicity with CF generation. \texttt{CF-Detox}$_{\text{tigtec}}$ enables to find a new compromise in terms of toxicity lowering, content preservation, and textual plausibility. We also showed that CFI could be used to enhance detoxification in terms of proximity and sparsity by following a \textit{target-then-replace-then-refine} approach. We believe that the latter concerns the XAI community in general, to make CF examples closer to the instance to be explained and sparser.

CF toxicity mitigation is highly dependent on the $f$ toxicity classifier used to steer detoxification. If $f$ is unreliable and only performs well on its training set, the risk of incorrectly indicating that the text has been detoxified is high. Therefore, the choice of~$f$ and the toxicity training data must be made with caution to avoid incorrect toxicity assessments during detoxification. The $f$ classifier can be fine-tuned on a more specific dataset if the detoxification task is related to a precise kind of toxicity, such as racism or sexism. 

CF detoxification has been tested by applying three different versions of \texttt{TIGTEC} with KernelSHAP, Integrated gradients, or Self-attention. There is a wide range of other LFI methods that could be used to target important tokens to explain a toxicity prediction. Besides, other CF generators such as \texttt{MiCE}~\cite{ross_explaining_2021}, \texttt{CREST}~\cite{treviso_crest_2023} or \texttt{CLOSS}~\cite{fern_text_2021} could be used to perform toxicity mitigation. We believe that these CF generation methods could lead to other levels of compromise between toxicity lowering, text plausibility and content preservation. 

Online toxicity is a systemic problem with complex and multiple roots~\cite{anatomy_online_hate}. Automatic toxicity processing does not in itself solve the factors \textit{causing} online toxic content generation, but they offer technological means of \textit{adaptation} to its rapid development. While CF toxicity mitigation may yield competitive results, it also raises ethical concerns that we discuss in Section~\ref{sec:use_misuse}.

\section{Ethical Considerations}
\label{sec:use_misuse}
 
The use of automatic processing tools raises crucial ethical considerations. We identify some of the risks associated with the use of toxicity mitigation tools, and draw recommendations to limit them.

\subsection{Managing  Diversity of Values}
In this work, we have considered a usual hate speech definition as "\textit{aggressive or offensive language that can be focused on a specific group of people who share common property such as religion, race, gender, sexual orientation, sex or political affiliation}". This definition is just one of many used by institutions and platforms to characterize hate speech~\cite{toxicity_survey_detection}. In particular, hate speech characterization can focus either on violence and hate incentives or on the objective of directly attacking. The choice of a specific definition can have a tangible impact on the way online toxicity is automatically processed and perceived by web users. In addition, the perceived toxicity of language can vary based on identity and beliefs~\cite{toxicity_divergence_belief_1}. For instance, conservative annotators can show a higher propensity to label African American English dialect as toxic while being less likely to annotate anti-Black comments as harmful~\cite{toxicity_divergence_belief_2}. Such annotator biases can be reflected in the datasets used to detect toxicity. \textbf{Recommendation}: Toxicity training sets have to be built by annotators carefully selected to represent diversity of values. Besides, toxicity training sets have to be chosen in order to to make sure that the values implicitly encoded in the classifier match those expected.


\subsection{Against Malicious Use}

A toxicity mitigation tool can be misused in various ways. One example is to use it as an adversarial attack generator to make a toxic text seem non-toxic. Adversarial attacks are small perturbations of data instances fooling a classifier with imperceptible changes~\cite{adversarial_attack}, which bring them formally close to CFs. Since toxicity mitigation methods are based on the use of a toxicity classifier to steer the detoxification process, these methods are subject to adversarial attacks. In this manner, a dishonest user could hijack a detoxification tool to find small modifications to the initial texts, leading a toxicity classifier to falsely assess that a text is correctly detoxified. \textbf{Recommendation}: A recent work~\cite{colombo2023toward} proposes a method to detect textual adversarial attacks based on the computation of similarities between an input embedding and the training distribution. Toxicity mitigation methods can also be reverse engineered to discover the rules used by an online platform to process harmful contents. This can lead to a change in the terms used, expressed in a seemingly neutral way in order to continue to express hateful content online~\cite{toxicity_creative}. \textbf{Recommendation}: Toxicity classifiers have to be frequently retrained on updated datasets integrating changes in the vocabulary used to express toxicity to overcome toxicity concept drift.


\subsection{Favoring Human-in-the-loop to Overcome Detoxification Inaccuracy}
State-of-the-art toxicity mitigation algorithms do not remove toxicity perfectly. Therefore, deploying a toxicity mitigation algorithm fully automatically is particularly risky since it could let harmful content spread. We propose to use these tools as the first layer of hate content processing before integrating humans into the loop. Behind any platform, online content must be reviewed and moderation is partly performed by human labor~\cite{custodian_internet, content_moderation_social_media}. By being exposed to disturbing toxic content, human moderators can develop psychological and emotional distress~\cite{psychological}.
Toxicity mitigation tools have the potential to induce a socio-technical change, suggesting textual intervention and reducing exposure for content moderators. \textbf{Recommendation}: We suggest using hate content detectors and mitigation methods based on the level of toxicity of the text. Text with the highest level of toxicity could simply be deleted, as it would be unlikely to be modified without completely altering its original meaning. Intermediate toxicity levels could be handled by a toxicity mitigation algorithm in order to preserve the general meaning of the text while proposing a softened version. This way, content moderators' exposure to the most hateful content would be significantly limited, and the more ambiguous content would be preprocessed by the mitigation algorithm to propose a more acceptable first version.

\subsection{Personal Data}
Each voluntary participant to the human evaluation signed an informed consent form outlining the project’s purpose and details. The data were anonymized and processed only by the authors and stored in accordance with the General Data Protection Regulation (GDPR). It was possible to stop at any time. The Consent form used is anonymized and presented in Appendix~\ref{sec:appendix}. This study was conducted for research purposes only.

\section{Conclusion}
This paper formalized how LFI and CF generation methods and CFI can be leveraged to accurately target textual toxic content and perform toxicity mitigation. \texttt{CF-Detox}$_{\text{tigtec}}$ leads to competitive results, with state-of-the-art performance in terms of content preservation while accurately detoxifying text and generating plausible text. \texttt{CF-Detox}$_{\text{tigtec}}$ is versatile since it can be used with various types of LFI methods (such as attention, gradient and perturbation) to target toxic POS and any kind of classifier allowing LFI computation. 

This paper is the first to show the extent to which fields such as automatic toxicity processing and explainable AI, which have developed in parallel, actually share epistemic similarities and can be mutually beneficial, paving the way towards practical applications of XAI methods.  
%
%
%
\bibliographystyle{plainnat}
\bibliography{detox}

\begin{thebibliography}{50}
\providecommand{\natexlab}[1]{#1}
\providecommand{\url}[1]{\texttt{#1}}
\expandafter\ifx\csname urlstyle\endcsname\relax
  \providecommand{\doi}[1]{doi: #1}\else
  \providecommand{\doi}{doi: \begingroup \urlstyle{rm}\Url}\fi

\bibitem[Al~Kuwatly et~al.(2020)Al~Kuwatly, Wich, and Groh]{toxicity_divergence_belief_1}
Hala Al~Kuwatly, Maximilian Wich, and Georg Groh.
\newblock Identifying and measuring annotator bias based on annotators{'} demographic characteristics.
\newblock In \emph{Proc. of the Fourth Workshop on Online Abuse and Harms}, pages 184--190. Association for Computational Linguistics, November 2020.
\newblock \doi{10.18653/v1/2020.alw-1.21}.
\newblock URL \url{https://aclanthology.org/2020.alw-1.21}.

\bibitem[Ali et~al.(2023)Ali, Abuhmed, El-Sappagh, Muhammad, Alonso-Moral, Confalonieri, Guidotti, Ser, Díaz-Rodríguez, and Herrera]{ali_explainable_2023}
Sajid Ali, Tamer Abuhmed, Shaker El-Sappagh, Khan Muhammad, Jose~M. Alonso-Moral, Roberto Confalonieri, Riccardo Guidotti, Javier~Del Ser, Natalia Díaz-Rodríguez, and Francisco Herrera.
\newblock Explainable {Artificial} {Intelligence} ({XAI}): {What} we know and what is left to attain {Trustworthy} {Artificial} {Intelligence}.
\newblock \emph{Information Fusion}, page 101805, April 2023.
\newblock ISSN 1566-2535.
\newblock \doi{10.1016/j.inffus.2023.101805}.
\newblock URL \url{https://www.sciencedirect.com/science/article/pii/S1566253523001148}.

\bibitem[Barredo~Arrieta et~al.(2020)Barredo~Arrieta, Díaz-Rodríguez, and Ser]{barredo_arrieta_explainable_2020}
Alejandro Barredo~Arrieta, Natalia Díaz-Rodríguez, and Del Ser.
\newblock Explainable {Artificial} {Intelligence} ({XAI}): {Concepts}, taxonomies, opportunities and challenges toward responsible {AI}.
\newblock \emph{Information Fusion}, 58:\penalty0 82--115, June 2020.
\newblock ISSN 1566-2535.
\newblock \doi{10.1016/j.inffus.2019.12.012}.
\newblock URL \url{https://www.sciencedirect.com/science/article/pii/S1566253519308103}.

\bibitem[Bender et~al.(2021)Bender, Gebru, McMillan-Major, and Shmitchell]{bender_dangers_2021}
Emily~M. Bender, Timnit Gebru, Angelina McMillan-Major, and Shmargaret Shmitchell.
\newblock On the {Dangers} of {Stochastic} {Parrots}: {Can} {Language} {Models} {Be} {Too} {Big}?
\newblock In \emph{Proc. of the 2021 {ACM} {Conf.} on {Fairness}, {Accountability}, and {Transparency}}, {FAccT} '21, pages 610--623, New York, NY, USA, March 2021. Association for Computing Machinery.
\newblock ISBN 978-1-4503-8309-7.
\newblock \doi{10.1145/3442188.3445922}.
\newblock URL \url{https://dl.acm.org/doi/10.1145/3442188.3445922}.

\bibitem[Bhan et~al.(2023{\natexlab{a}})Bhan, Achache, Legrand, Blangero, and Chesneau]{bhan_evaluating_2023}
Milan Bhan, Nina Achache, Victor Legrand, Annabelle Blangero, and Nicolas Chesneau.
\newblock Evaluating self-attention interpretability through human-grounded experimental protocol.
\newblock In \emph{Proc. of the First World Conf. on Explainable Artificial Intelligence xAI}, pages 26--46, 2023{\natexlab{a}}.
\newblock URL \url{http://arxiv.org/abs/2303.15190}.

\bibitem[Bhan et~al.(2023{\natexlab{b}})Bhan, Vittaut, Chesneau, and Lesot]{cfi}
Milan Bhan, Jean-noel Vittaut, Nicolas Chesneau, and Marie-jeanne Lesot.
\newblock Enhancing textual counterfactual explanation intelligibility through counterfactual feature importance.
\newblock In \emph{Proceedings of the 3rd Workshop on Trustworthy Natural Language Processing (TrustNLP 2023)}, pages 221--231, Toronto, Canada, July 2023{\natexlab{b}}. Association for Computational Linguistics.
\newblock \doi{10.18653/v1/2023.trustnlp-1.19}.
\newblock URL \url{https://aclanthology.org/2023.trustnlp-1.19}.

\bibitem[Bhan et~al.(2023{\natexlab{c}})Bhan, Vittaut, Chesneau, and Lesot]{tigtec_2}
Milan Bhan, Jean-Noel Vittaut, Nicolas Chesneau, and Marie-Jeanne Lesot.
\newblock Tigtec: Token importance guided text counterfactuals.
\newblock In \emph{Proc. of the European Conf. on Machine Learning ECML-PKDD}, page 496–512. Springer, 2023{\natexlab{c}}.

\bibitem[Brown et~al.(2020)Brown, Mann, Ryder, and Sutskever]{brown_language_2020}
Tom Brown, Benjamin Mann, Nick Ryder, and Ilya Sutskever.
\newblock Language {Models} are {Few}-{Shot} {Learners}.
\newblock In \emph{Advances in {Neural} {Information} {Processing} {Systems}}, volume~33, pages 1877--1901, 2020.

\bibitem[Casta{\~n}o-Pulgar{\'\i}n et~al.(2021)Casta{\~n}o-Pulgar{\'\i}n, Su{\'a}rez-Betancur, Vega, and L{\'o}pez]{castano2021internet}
Sergio~Andr{\'e}s Casta{\~n}o-Pulgar{\'\i}n, Natalia Su{\'a}rez-Betancur, Luz Magnolia~Tilano Vega, and Harvey Mauricio~Herrera L{\'o}pez.
\newblock Internet, social media and online hate speech. systematic review.
\newblock \emph{Aggression and violent behavior}, 58:\penalty0 101608, 2021.

\bibitem[Colombo et~al.(2023)Colombo, Picot, Noiry, Staerman, and Piantanida]{colombo2023toward}
Pierre Colombo, Marine Picot, Nathan Noiry, Guillaume Staerman, and Pablo Piantanida.
\newblock Toward stronger textual attack detectors.
\newblock In \emph{Findings of the Association for Computational Linguistics: EMNLP 2023}, pages 484--505, 2023.

\bibitem[Cook et~al.(2023)Cook, Zilka, DeSandre, Giles, and Maskell]{protecting_children}
Darren Cook, Miri Zilka, Heidi DeSandre, Susan Giles, and Simon Maskell.
\newblock Protecting children from online exploitation: Can a trained model detect harmful communication strategies?
\newblock In \emph{Proc. of the 2023 AAAI/ACM Conference on AI, Ethics, and Society}, AIES '23, page 5–14. Association for Computing Machinery, 2023.
\newblock \doi{10.1145/3600211.3604696}.
\newblock URL \url{https://doi.org/10.1145/3600211.3604696}.

\bibitem[Dale et~al.(2021)Dale, Voronov, and Dementieva]{dale_text_2021}
David Dale, Anton Voronov, and Daryna Dementieva.
\newblock Text {Detoxification} using {Large} {Pre}-trained {Neural} {Models}.
\newblock In Marie-Francine Moens, Xuanjing Huang, Lucia Specia, and Scott Wen-tau Yih, editors, \emph{Proc. of the 2021 {Conf.} on {Empirical} {Methods} in {Natural} {Language} {Processing}}, pages 7979--7996. Association for Computational Linguistics, November 2021.
\newblock \doi{10.18653/v1/2021.emnlp-main.629}.
\newblock URL \url{https://aclanthology.org/2021.emnlp-main.629}.

\bibitem[Fern and Pope(2021)]{fern_text_2021}
Xiaoli Fern and Quintin Pope.
\newblock Text {Counterfactuals} via {Latent} {Optimization} and {Shapley}-{Guided} {Search}.
\newblock In \emph{Proc. of the 2021 {Conf.} on {Empirical} {Methods} in {Natural} {Language} {Processing}}, pages 5578--5593. Association for Computational Linguistics, November 2021.
\newblock \doi{10.18653/v1/2021.emnlp-main.452}.
\newblock URL \url{https://aclanthology.org/2021.emnlp-main.452}.

\bibitem[Fortuna and Nunes(2018)]{toxicity_survey_detection}
Paula Fortuna and S\'{e}rgio Nunes.
\newblock A survey on automatic detection of hate speech in text.
\newblock \emph{ACM Comput. Surv.}, 51\penalty0 (4), jul 2018.
\newblock ISSN 0360-0300.
\newblock \doi{10.1145/3232676}.
\newblock URL \url{https://doi.org/10.1145/3232676}.

\bibitem[Gallegos et~al.(2023)Gallegos, Rossi, Barrow, Tanjim, Kim, Dernoncourt, Yu, Zhang, and Ahmed]{gallegos_bias_2023}
Isabel~O. Gallegos, Ryan~A. Rossi, Joe Barrow, Md~Mehrab Tanjim, Sungchul Kim, Franck Dernoncourt, Tong Yu, Ruiyi Zhang, and Nesreen~K. Ahmed.
\newblock Bias and {Fairness} in {Large} {Language} {Models}: {A} {Survey}, September 2023.
\newblock URL \url{http://arxiv.org/abs/2309.00770}.
\newblock arXiv:2309.00770 [cs].

\bibitem[Gemma~Team et~al.(2024)Gemma~Team, Mesnard, Hardin, Dadashi, et~al.]{team2024gemma}
Gemma Gemma~Team, Thomas Mesnard, Cassidy Hardin, Dadashi, et~al.
\newblock Gemma: Open models based on gemini research and technology.
\newblock \emph{arXiv preprint arXiv:2403.08295}, 2024.

\bibitem[Gillespie(2018)]{custodian_internet}
Tarleton Gillespie.
\newblock \emph{Custodians of the Internet: Platforms, content moderation, and the hidden decisions that shape social media}.
\newblock Yale University Press, 2018.

\bibitem[Guidotti(2022)]{guidotti_counterfactual_2022}
Riccardo Guidotti.
\newblock Counterfactual explanations and how to find them: literature review and benchmarking.
\newblock \emph{Data Mining and Knowledge Discovery}, April 2022.
\newblock ISSN 1573-756X.
\newblock \doi{10.1007/s10618-022-00831-6}.
\newblock URL \url{https://doi.org/10.1007/s10618-022-00831-6}.

\bibitem[Hallinan et~al.(2023)Hallinan, Liu, Choi, and Sap]{hallinan_detoxifying_2023}
Skyler Hallinan, Alisa Liu, Yejin Choi, and Maarten Sap.
\newblock Detoxifying {Text} with {MaRCo}: {Controllable} {Revision} with {Experts} and {Anti}-{Experts}.
\newblock In \emph{Proc. of the 61st {Annual} {Meeting} of the {Association} for {Computational} {Linguistics} ({Volume} 2: {Short} {Papers})}, pages 228--242. Association for Computational Linguistics, July 2023.
\newblock URL \url{https://aclanthology.org/2023.acl-short.21}.

\bibitem[Hu et~al.(2022)Hu, Lee, Aggarwal, and Zhang]{tst_review}
Zhiqiang Hu, Roy Ka-Wei Lee, Charu~C. Aggarwal, and Aston Zhang.
\newblock Text style transfer: A review and experimental evaluation.
\newblock \emph{SIGKDD Explor. Newsl.}, 24\penalty0 (1):\penalty0 14–45, jun 2022.
\newblock ISSN 1931-0145.
\newblock \doi{10.1145/3544903.3544906}.
\newblock URL \url{https://doi.org/10.1145/3544903.3544906}.

\bibitem[Jelinek et~al.(2005)Jelinek, Mercer, Bahl, and Baker]{jelinek_perplexitymeasure_2005}
F.~Jelinek, R.~L. Mercer, L.~R. Bahl, and J.~K. Baker.
\newblock Perplexity—a measure of the difficulty of speech recognition tasks.
\newblock \emph{The Journal of the Acoustical Society of America}, 62\penalty0 (S1):\penalty0 S63, August 2005.
\newblock ISSN 0001-4966.
\newblock \doi{10.1121/1.2016299}.
\newblock URL \url{https://doi.org/10.1121/1.2016299}.

\bibitem[Jhaver et~al.(2023)Jhaver, Zhang, Chen, Natarajan, Wang, and Zhang]{content_moderation_social_media}
Shagun Jhaver, Alice~Qian Zhang, Quan~Ze Chen, Nikhila Natarajan, Ruotong Wang, and Amy~X. Zhang.
\newblock Personalizing content moderation on social media: User perspectives on moderation choices, interface design, and labor.
\newblock \emph{Proc. ACM Hum.-Comput. Interact.}, 7\penalty0 (CSCW2), oct 2023.
\newblock \doi{10.1145/3610080}.
\newblock URL \url{https://doi.org/10.1145/3610080}.

\bibitem[Ji and Knight(2018)]{toxicity_creative}
Heng Ji and Kevin Knight.
\newblock Creative language encoding under censorship.
\newblock In Chris Brew, Anna Feldman, and Chris Leberknight, editors, \emph{Proc. of the First Workshop on Natural Language Processing for {I}nternet Freedom}, pages 23--33. Association for Computational Linguistics, August 2018.
\newblock URL \url{https://aclanthology.org/W18-4203}.

\bibitem[Jigsaw(2018)]{jigsaw_18}
Jigsaw.
\newblock Toxic comment classification challenge.
\newblock \url{https://www.kaggle.com/c/jigsaw-toxic-comment-classification-challenge}, 2018.
\newblock Accessed: 2010-09-30.

\bibitem[Laugel et~al.(2019)Laugel, Lesot, Marsala, Renard, and Detyniecki]{laugel_dangers_2019}
Thibault Laugel, Marie-Jeanne Lesot, Christophe Marsala, Xavier Renard, and Marcin Detyniecki.
\newblock The {Dangers} of {Post}-hoc {Interpretability}: {Unjustified} {Counterfactual} {Explanations}.
\newblock In \emph{Proc. of the {Twenty}-{Eighth} {Int.} {Joint} {Conf.} on {Artificial} {Intelligence}}, pages 2801--2807. Int. Joint Conf. on Artificial Intelligence Organization, August 2019.
\newblock ISBN 978-0-9992411-4-1.
\newblock \doi{10.24963/ijcai.2019/388}.
\newblock URL \url{https://www.ijcai.org/proceedings/2019/388}.

\bibitem[Laugier et~al.(2021)Laugier, Pavlopoulos, Sorensen, and Dixon]{laugier2021civil}
L{\'e}o Laugier, John Pavlopoulos, Jeffrey Sorensen, and Lucas Dixon.
\newblock Civil rephrases of toxic texts with self-supervised transformers.
\newblock In \emph{Proc. of the 16th Conference of the European Chapter of the Association for Computational Linguistics: Main Volume}, pages 1442--1461, 2021.

\bibitem[Lewis et~al.(2020)Lewis, Liu, Goyal, Ghazvininejad, Mohamed, Levy, Stoyanov, and Zettlemoyer]{lewis_bart_2020}
Mike Lewis, Yinhan Liu, Naman Goyal, Marjan Ghazvininejad, Abdelrahman Mohamed, Omer Levy, Veselin Stoyanov, and Luke Zettlemoyer.
\newblock {BART}: {Denoising} {Sequence}-to-{Sequence} {Pre}-training for {Natural} {Language} {Generation}, {Translation}, and {Comprehension}.
\newblock In \emph{Proc. of the 58th {Annual} {Meeting} of the {Association} for {Computational} {Linguistics}}, pages 7871--7880, Online, July 2020. Association for Computational Linguistics.
\newblock \doi{10.18653/v1/2020.acl-main.703}.
\newblock URL \url{https://aclanthology.org/2020.acl-main.703}.

\bibitem[Liang et~al.(2022)Liang, He, Zhao, Jia, and Li]{adversarial_attack}
Hongshuo Liang, Erlu He, Yangyang Zhao, Zhe Jia, and Hao Li.
\newblock Adversarial attack and defense: A survey.
\newblock \emph{Electronics}, 11\penalty0 (8):\penalty0 1283, 2022.

\bibitem[Lundberg and Lee(2017)]{lundberg_unified_2017}
Scott~M. Lundberg and Su-In Lee.
\newblock A unified approach to interpreting model predictions.
\newblock In \emph{Proc. of the 31st {Int.} {Conf.} on {Neural} {Information} {Processing} {Systems}}, {NIPS}'17, pages 4768--4777, December 2017.
\newblock ISBN 978-1-5108-6096-4.

\bibitem[Madaan et~al.(2022)Madaan, Bedathur, and Saha]{madaan_plug_2022}
Nishtha Madaan, Srikanta Bedathur, and Diptikalyan Saha.
\newblock Plug and {Play} {Counterfactual} {Text} {Generation} for {Model} {Robustness}, June 2022.
\newblock URL \url{http://arxiv.org/abs/2206.10429}.
\newblock arXiv:2206.10429 [cs].

\bibitem[Mart{\'\i}nez et~al.(2023)Mart{\'\i}nez, Watson, Reviriego, Hern{\'a}ndez, Juarez, and Sarkar]{gen_ai_internet}
Gonzalo Mart{\'\i}nez, Lauren Watson, Pedro Reviriego, Jos{\'e}~Alberto Hern{\'a}ndez, Marc Juarez, and Rik Sarkar.
\newblock Combining generative artificial intelligence (ai) and the internet: Heading towards evolution or degradation?
\newblock \emph{arXiv preprint arXiv:2303.01255}, 2023.

\bibitem[Miller(2019)]{miller_explanation_2019}
Tim Miller.
\newblock Explanation in artificial intelligence: {Insights} from the social sciences.
\newblock \emph{Artificial Intelligence}, 267:\penalty0 1--38, February 2019.
\newblock ISSN 0004-3702.
\newblock \doi{10.1016/j.artint.2018.07.007}.
\newblock URL \url{https://www.sciencedirect.com/science/article/pii/S0004370218305988}.

\bibitem[Molnar(2020)]{molnar_interpretable_2020}
Christoph Molnar.
\newblock \emph{Interpretable {Machine} {Learning}}.
\newblock Lulu.com, 2020.
\newblock ISBN 978-0-244-76852-2.
\newblock URL \url{https://christophm.github.io/interpretable-ml-book/}.

\bibitem[Nogueira~dos Santos et~al.(2018)Nogueira~dos Santos, Melnyk, and Padhi]{detox_style_transfer}
Cicero Nogueira~dos Santos, Igor Melnyk, and Inkit Padhi.
\newblock Fighting offensive language on social media with unsupervised text style transfer.
\newblock In Iryna Gurevych and Yusuke Miyao, editors, \emph{Proc. of the 56th Annual Meeting of the Association for Computational Linguistics (Volume 2: Short Papers)}, pages 189--194. Association for Computational Linguistics, July 2018.
\newblock \doi{10.18653/v1/P18-2031}.
\newblock URL \url{https://aclanthology.org/P18-2031}.

\bibitem[Papineni et~al.(2002)Papineni, Roukos, Ward, and Zhu]{BLEU}
Kishore Papineni, Salim Roukos, Todd Ward, and Wei-Jing Zhu.
\newblock Bleu: a method for automatic evaluation of machine translation.
\newblock In \emph{Proc. of the 40th Annual Meeting on Association for Computational Linguistics}, ACL '02, page 311–318, 2002.
\newblock \doi{10.3115/1073083.1073135}.
\newblock URL \url{https://doi.org/10.3115/1073083.1073135}.

\bibitem[Pennington et~al.(2014)Pennington, Socher, and Manning]{glove}
Jeffrey Pennington, Richard Socher, and Christopher Manning.
\newblock {G}lo{V}e: Global vectors for word representation.
\newblock In \emph{Proc. of the 2014 Conf. on Empirical Methods in Natural Language Processing ({EMNLP})}, pages 1532--1543. Association for Computational Linguistics, October 2014.
\newblock \doi{10.3115/v1/D14-1162}.
\newblock URL \url{https://aclanthology.org/D14-1162}.

\bibitem[Radford et~al.(2018)Radford, Wu, Child, Luan, Amodei, and Sutskever]{radford_language_nodate}
Alec Radford, Jeffrey Wu, Rewon Child, David Luan, Dario Amodei, and Ilya Sutskever.
\newblock Language {Models} are {Unsupervised} {Multitask} {Learners}.
\newblock 2018.

\bibitem[Rapp(2021)]{call_to_genocide}
Kyle Rapp.
\newblock Social media and genocide: The case for home state responsibility.
\newblock \emph{Journal of Human Rights}, 20\penalty0 (4):\penalty0 486--502, 2021.

\bibitem[Raﬀel et~al.(2019)Raﬀel, Shazeer, Roberts, Lee, Narang, Matena, Zhou, Li, and Liu]{rael_exploring_nodate}
Colin Raﬀel, Noam Shazeer, Adam Roberts, Katherine Lee, Sharan Narang, Michael Matena, Yanqi Zhou, Wei Li, and Peter~J Liu.
\newblock Exploring the {Limits} of {Transfer} {Learning} with a {Uniﬁed} {Text}-to-{Text} {Transformer}.
\newblock 2019.

\bibitem[Reimers and Gurevych(2019)]{reimers_sentence-bert_2019}
Nils Reimers and Iryna Gurevych.
\newblock Sentence-{BERT}: {Sentence} {Embeddings} using {Siamese} {BERT}-{Networks}.
\newblock In \emph{Proc. of the 2019 {Conf.} on {Empirical} {Methods} in {Natural} {Language} {Processing} and the 9th {Int.} {Joint} {Conf.} on {Natural} {Language} {Processing} ({EMNLP}-{IJCNLP})}, pages 3982--3992. Association for Computational Linguistics, November 2019.
\newblock \doi{10.18653/v1/D19-1410}.
\newblock URL \url{https://aclanthology.org/D19-1410}.

\bibitem[Ross et~al.(2021)Ross, Marasović, and Peters]{ross_explaining_2021}
Alexis Ross, Ana Marasović, and Matthew Peters.
\newblock Explaining {NLP} {Models} via {Minimal} {Contrastive} {Editing} ({MiCE}).
\newblock In \emph{Findings of the {Association} for {Computational} {Linguistics}: {ACL}-{IJCNLP} 2021}, pages 3840--3852, Online, August 2021. Association for Computational Linguistics.
\newblock \doi{10.18653/v1/2021.findings-acl.336}.
\newblock URL \url{https://aclanthology.org/2021.findings-acl.336}.

\bibitem[Salminen et~al.(2018)Salminen, Almerekhi, Milenkovi{\'c}, Jung, An, and Kwak]{anatomy_online_hate}
Joni Salminen, Hind Almerekhi, Milica Milenkovi{\'c}, Soon-gyo Jung, Jisun An, and Haewoon Kwak.
\newblock Anatomy of online hate: developing a taxonomy and machine learning models for identifying and classifying hate in online news media.
\newblock In \emph{Proc. of the Int. AAAI Conf. on Web and Social Media}, volume~12, 2018.

\bibitem[Sap et~al.(2022)Sap, Swayamdipta, Vianna, Zhou, Choi, and Smith]{toxicity_divergence_belief_2}
Maarten Sap, Swabha Swayamdipta, Laura Vianna, Xuhui Zhou, Yejin Choi, and Noah~A. Smith.
\newblock Annotators with attitudes: How annotator beliefs and identities bias toxic language detection.
\newblock In \emph{Proc. of the 2022 Conf. of the North American Chapter of the Association for Computational Linguistics: Human Language Technologies}, pages 5884--5906, Seattle, United States, July 2022. Association for Computational Linguistics.
\newblock \doi{10.18653/v1/2022.naacl-main.431}.
\newblock URL \url{https://aclanthology.org/2022.naacl-main.431}.

\bibitem[Spence et~al.(2023)Spence, Bifulco, Bradbury, Martellozzo, and DeMarco]{psychological}
Ruth Spence, Antonia Bifulco, Paula Bradbury, Elena Martellozzo, and Jeffrey DeMarco.
\newblock The psychological impacts of content moderation on content moderators: A qualitative study.
\newblock \emph{Cyberpsychology: Journal of Psychosocial Research on Cyberspace}, 17\penalty0 (4), 2023.

\bibitem[Sundararajan et~al.(2017)Sundararajan, Taly, and Yan]{sundararajan_axiomatic_2017}
Mukund Sundararajan, Ankur Taly, and Qiqi Yan.
\newblock Axiomatic attribution for deep networks.
\newblock In \emph{Proc. of the 34th {Int.} {Conf.} on {Machine} {Learning}, ICML}, volume~70 of \emph{{ICML}'17}, pages 3319--3328. JMLR.org, August 2017.
\newblock URL \url{https://proceedings.mlr.press/v70/sundararajan17a/sundararajan17a.pdf}.

\bibitem[Thomas et~al.(2021)Thomas, Akhawe, Bailey, Boneh, Bursztein, and Consolvo]{hate_online}
Kurt Thomas, Devdatta Akhawe, Michael Bailey, Dan Boneh, Elie Bursztein, and Sunny Consolvo.
\newblock Sok: Hate, harassment, and the changing landscape of online abuse.
\newblock In \emph{2021 IEEE Symposium on Security and Privacy (SP)}, pages 247--267, 2021.
\newblock \doi{10.1109/SP40001.2021.00028}.

\bibitem[Tran et~al.(2020)Tran, Zhang, and Soleymani]{detox_style_transfer_2}
Minh Tran, Yipeng Zhang, and Mohammad Soleymani.
\newblock Towards a friendly online community: An unsupervised style transfer framework for profanity redaction.
\newblock In Donia Scott, Nuria Bel, and Chengqing Zong, editors, \emph{Proc. of the 28th Int. Conf. on Computational Linguistics}, pages 2107--2114. International Committee on Computational Linguistics, December 2020.
\newblock \doi{10.18653/v1/2020.coling-main.190}.
\newblock URL \url{https://aclanthology.org/2020.coling-main.190}.

\bibitem[Treviso et~al.(2023)Treviso, Ross, Guerreiro, and Martins]{treviso_crest_2023}
Marcos Treviso, Alexis Ross, Nuno~M. Guerreiro, and Andr{\'e} Martins.
\newblock {CREST}: A joint framework for rationalization and counterfactual text generation.
\newblock In Anna Rogers, Jordan Boyd-Graber, and Naoaki Okazaki, editors, \emph{Proc. of the 61st Annual Meeting of the Association for Computational Linguistics (Volume 1: Long Papers)}, pages 15109--15126. Association for Computational Linguistics, July 2023.
\newblock \doi{10.18653/v1/2023.acl-long.842}.
\newblock URL \url{https://aclanthology.org/2023.acl-long.842}.

\bibitem[Walther(2022)]{social_media_hate_impact}
Joseph Walther.
\newblock Social media and online hate.
\newblock \emph{Current Opinion in Psychology}, 45, 01 2022.
\newblock \doi{10.1016/j.copsyc.2021.12.010}.

\bibitem[Zhang et~al.(2019)Zhang, Kishore, Wu, Weinberger, and Artzi]{zhang2019bertscore}
Tianyi Zhang, Varsha Kishore, Felix Wu, Kilian~Q Weinberger, and Yoav Artzi.
\newblock Bertscore: Evaluating text generation with bert.
\newblock In \emph{Int. Conf. on Learning Representations}, 2019.

\end{thebibliography}
%




\newpage
\appendix
\section{Appendix}
\label{sec:appendix}

\subsection{Consent form}
\begin{figure}
    \centering
    {\includegraphics[width=0.4\textwidth]{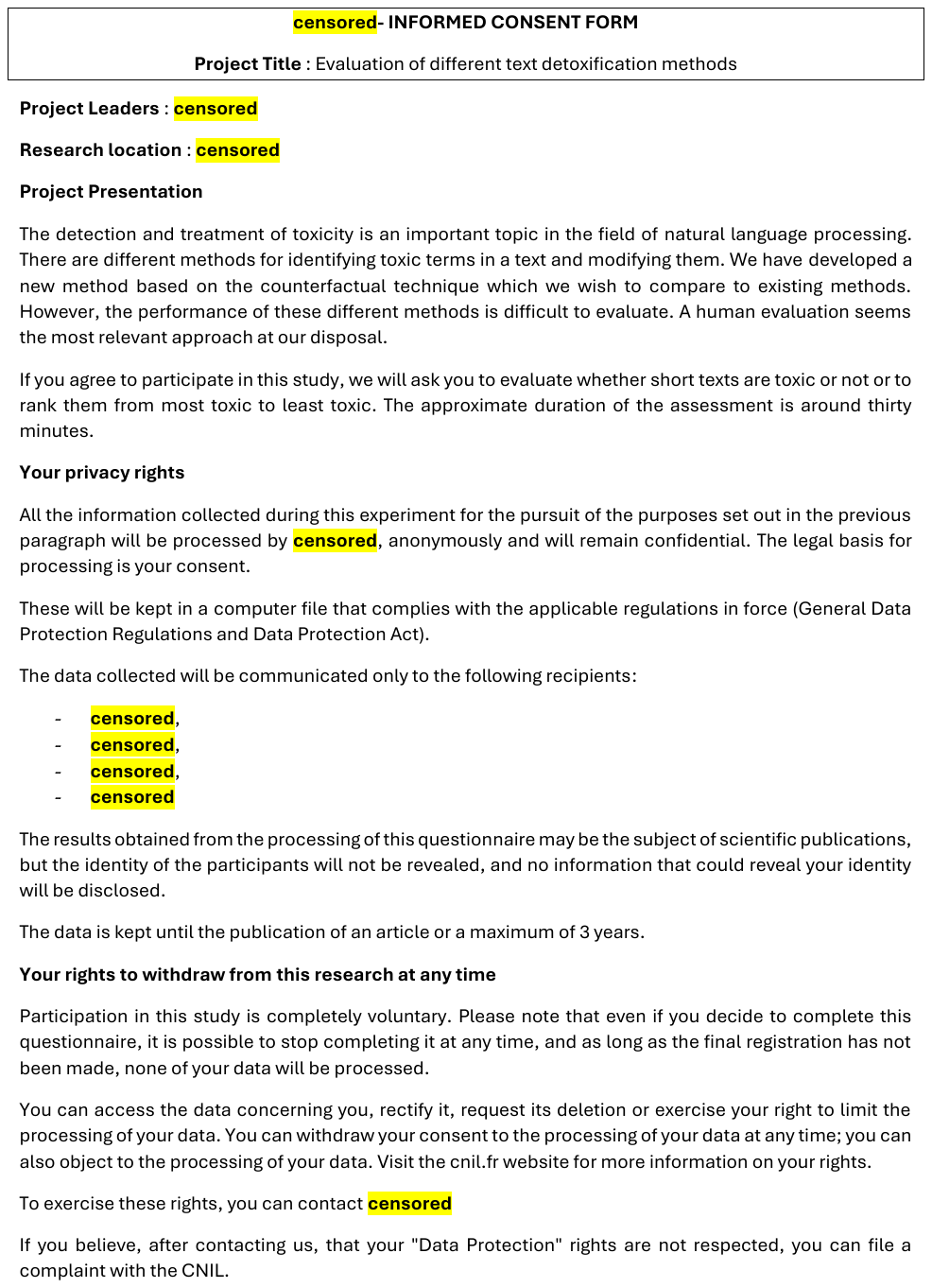}} 
    {\includegraphics[width=0.4\textwidth]{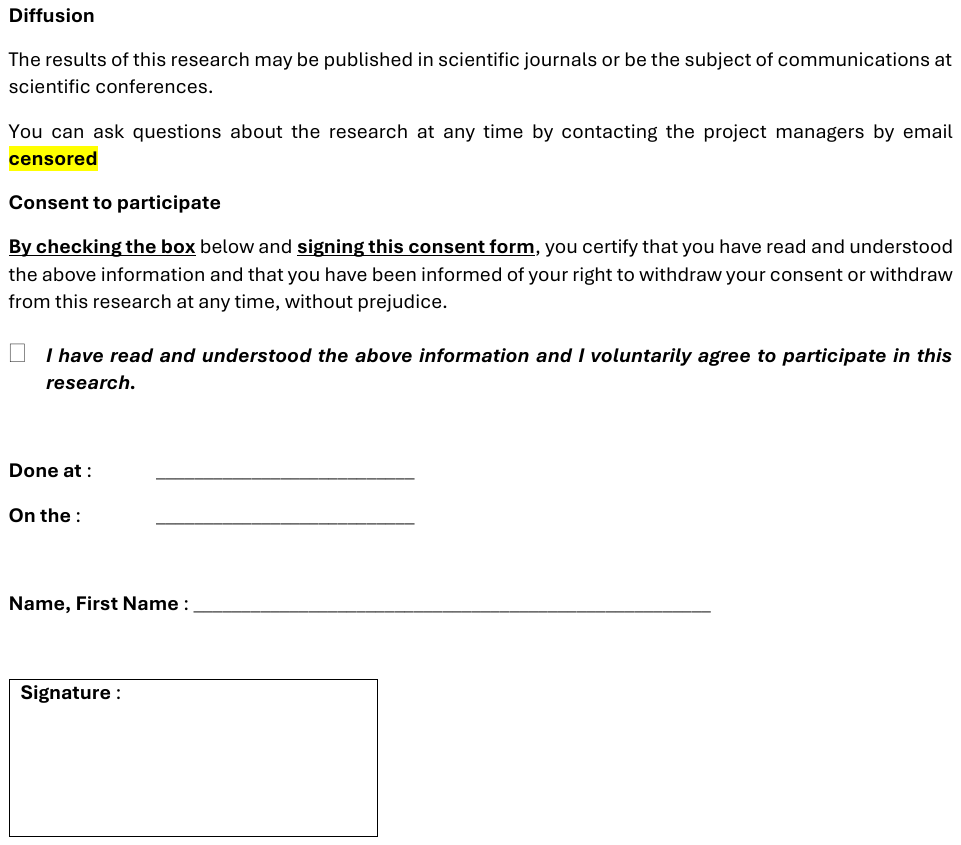}} 
    \caption{\label{fig:consent_form_1} Consent form}
    \label{fig:foobar}
\end{figure}


\subsection{Implementation Details}
\label{sec:appendix_slm_implementation_details}
The code used to run the experiments is available on the github related to the TIGTEC paper\footnote{\url{https://github.com/milanbhan/tigtec}}.
\paragraph{Pretrained language models} The library used to import pretrained language model to be finetuned to distinguish toxic and non-toxic content is \verb|transformer|. In particular, the backbone version of BERT is \texttt{bert-base-uncased} the one of DistilBERT is \texttt{distilbert-base-uncased}. The library used to import the Sentence Transformer is \verb|sentence_transformers| and the model backbone is \verb|paraphrase-MiniLM-L6-v2|. The backbone of the pre-trained toxicity classifier is \verb|toxic-bert|.
\end{document}